\documentclass[preprint,12pt]{elsarticle}

\usepackage[font=small,labelsep=period]{caption}
\usepackage{amssymb}
\usepackage{amsmath,amsfonts}
\usepackage{algorithm}
\usepackage{array}
\usepackage{textcomp}
\usepackage{stfloats}
\usepackage{url}
\usepackage{verbatim}
\usepackage{graphicx}
\usepackage{subfigure}
\usepackage{xcolor}
\usepackage{multirow}

\usepackage{rotating}
\usepackage{pdflscape}

\begin{document}

\begin{frontmatter}



\title{Heuristics for Vehicle Routing Problem: A Survey and Recent Advances}


\author[a]{Fei Liu\corref{cor1}}
\ead{fliu36-c@my.cityu.edu.hk}
\cortext[cor1]{Corresponding author}
\author[a]{Chengyu Lu}
\author[a,b]{Lin Gui}
\author[a]{Qingfu Zhang}
\author[c]{Xialiang Tong}
\author[c]{Mingxuan Yuan}

\affiliation[a]{organization={Computer Science, City Univeristy of Hong Kong},
            country={Hong Kong}}

\affiliation[b]{organization={State Key Laboratory of Digital Manufacturing Equipment and Technology, Huazhong University of Science and Technology},
            city={Wuhan}, 
            country={China}}
\affiliation[c]{organization={Huawei Noah’s Ark Lab},
            country={Hong Kong}}

\begin{abstract}
Vehicle routing is a well-known optimization research topic with significant practical importance. Among different approaches to solving vehicle routing, heuristics can produce a satisfactory solution at a reasonable computational cost. Consequently, much effort has been made in the past decades to develop vehicle routing heuristics. In this article, we systematically survey the existing vehicle routing heuristics, particularly on works carried out in recent years. A classification of vehicle routing heuristics is presented, followed by a review of their methodologies, recent developments, and applications. Moreover, we present a general framework of state-of-the-art methods and provide insights into their success. Finally, three emerging research topics with notable works and future directions are discussed. 
\end{abstract}




\end{frontmatter}



\section{Introduction}\label{sec1}

Vehicle routing problem (VRP) can be found in many real-life applications such as logistics, transportation, manufacturing, retail distribution, waste collection, and delivery planning~\cite{toth2014vehicle}. It is about managing a fleet of vehicles to serve the requests of a set of customers and minimize the routing cost. The first VRP was proposed by~\cite{dantzig1959truck} in 1959. With the effort of more than six decades, a large number of VRP variants with real-world attributes (e.g., time windows, heterogeneous fleet, and multiple depots) have been studied~\cite{tan2021vehicle}. The number of publications on vehicle routing is growing dramatically at a rate of six percent per year as is noted in~\cite{eksioglu2009vehicle}. The interest is not only motivated by the rich structure and difficulty as an optimization problem but also by its practical significance.

Among different approaches for solving vehicle routing problems, exact methods~\cite{baldacci2012recent,costa2019exact} have a guarantee of solution optimality and are more theoretically sound. However, it is difficult for them to handle the increasingly complex modern real-world vehicle routing problems efficiently. Neural combinatorial optimization~\cite{nazari2018reinforcement,bengio2021machine,kotary2021end,wang2023solving}, as an emerging trend, generates an acceptable solution quickly in an end-to-end manner with the help of learned knowledge from a large amount of data. However, its results can hardly beat cutting-edge heuristics. In contrast, heuristics provide a high-quality solution at a reasonable computational cost and have good generalization ability, and thus have gained popularity among both researchers and practitioners~\cite{weinand2022research}. To this end, much effort has been made in the past decades to develop vehicle routing heuristics~\cite{gendreau2008metaheuristics,elshaer2020taxonomic}. According to~\cite{braekers2016vehicle}, heuristics made up more than 80\% of the vehicle routing literature published between 2009 and 2013.

\begin{figure}[htbp]
    \centering  
    \includegraphics[width=1.0\textwidth]{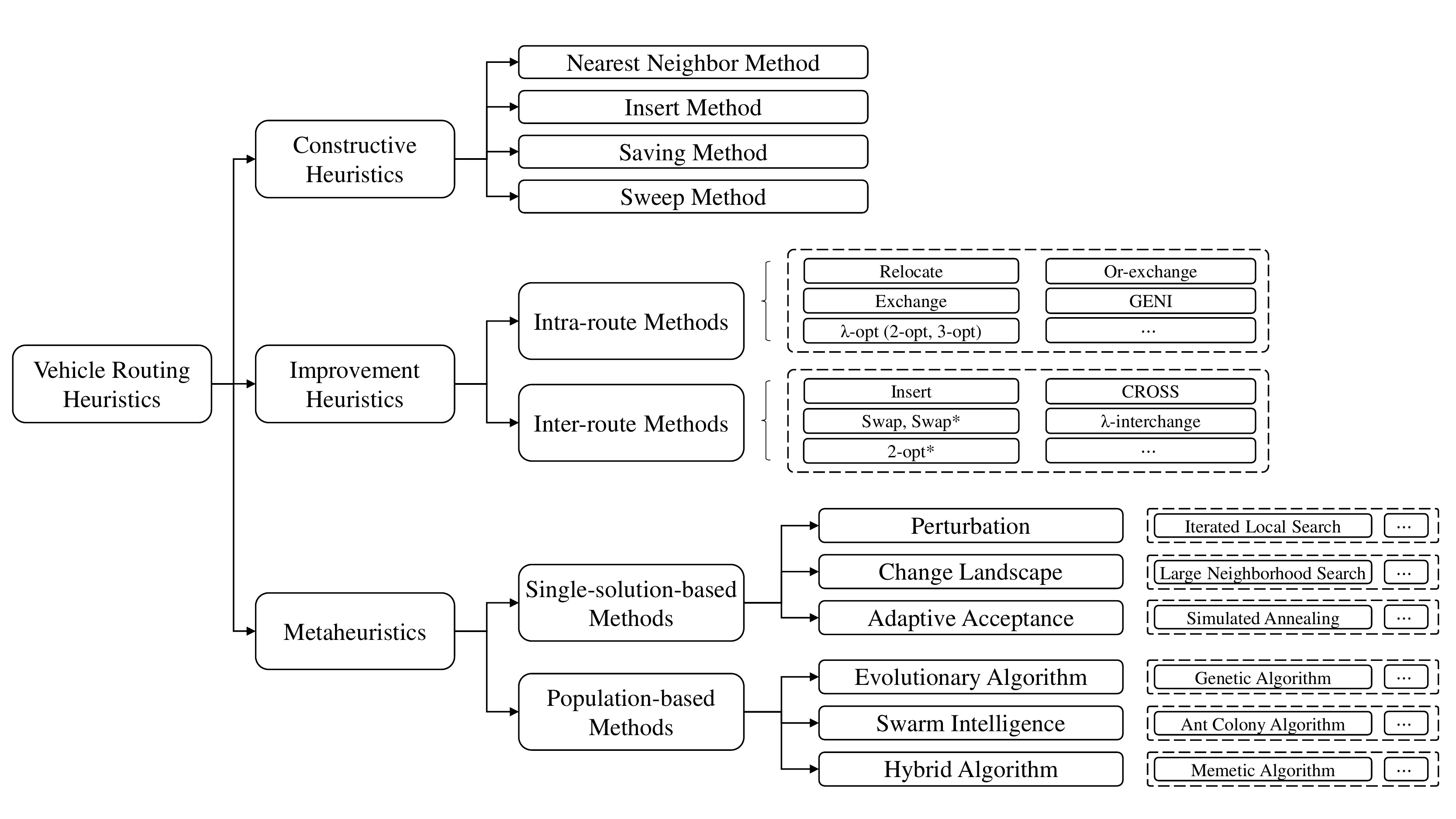}
    \caption{A classification of vehicle routing heuristics}
    \label{fig:heuristic_class}
\end{figure}

The research on vehicle routing heuristics appeared as early as the VRP itself~\cite{dantzig1959truck}. As illustrated in Fig.~\ref{fig:heuristic_class}, we present a classification of vehicle routing heuristics to introduce the extensive work in this area. The heuristics are categorized into three main categories: 
\begin{enumerate}
    \item Constructive heuristics: The algorithms in this category build routing solutions from zero following some fixed empirical heuristic procedures. They usually generate a feasible solution fast and are easy to implement on different VRP variants. The solution produced by constructive heuristics, however, often has a certain gap to the optimal solution~\cite{laporte2002classical,kosasih2020comparison}. 
    \item Improvement heuristics: They iteratively improve an incumbent routing solution by performing a local search in the neighborhood. They are quite efficient in determining a local optimum. The main limitation is that they can be easily trapped in local optimal and the final solution's quality depends on the start point of the local search~\cite{laporte2002classical,toth2014vehicle}. 
    \item Metaheuristics: Different from constructive heuristics and improvement heuristics, which try to take full advantage of the feature and structure of the problem, metaheuristics provide high-level algorithm principles~\cite{gendreau2010handbook} and are less problem-dependent. They typically bring ideas from natural phenomena or physical processes to design the optimization algorithm paradigm. They are usually efficient and have global search ability.
\end{enumerate}

Several works have been produced to summarize the achievements of vehicle routing research studies. They can be found in papers~\cite{laporte1992vehicle,cordeau2005new,prodhon2016metaheuristics,goel2017vehicle,eksioglu2009vehicle,braekers2016vehicle,rasku2019meta,elshaer2020taxonomic,konstantakopoulos2022vehicle,tan2021vehicle} and book chapters~\cite{cordeau2007vehicle,toth2014vehicle,labadie2016metaheuristics}. However, very few of them were dedicated to a general survey of vehicle routing heuristics. To the best of our knowledge, the most related works are~\cite{cordeau2002guide} and~\cite{vidal2013heuristics}. The former provided a comprehensive guide on heuristics for vehicle routing problems and compared different heuristics concerning four performance measurement criteria. It was presented two decades ago. The latter reviewed the works from the perspective of multi-attribute VRPs.

\begin{figure}[htbp]
    \centering
    \includegraphics[width=0.8\textwidth]{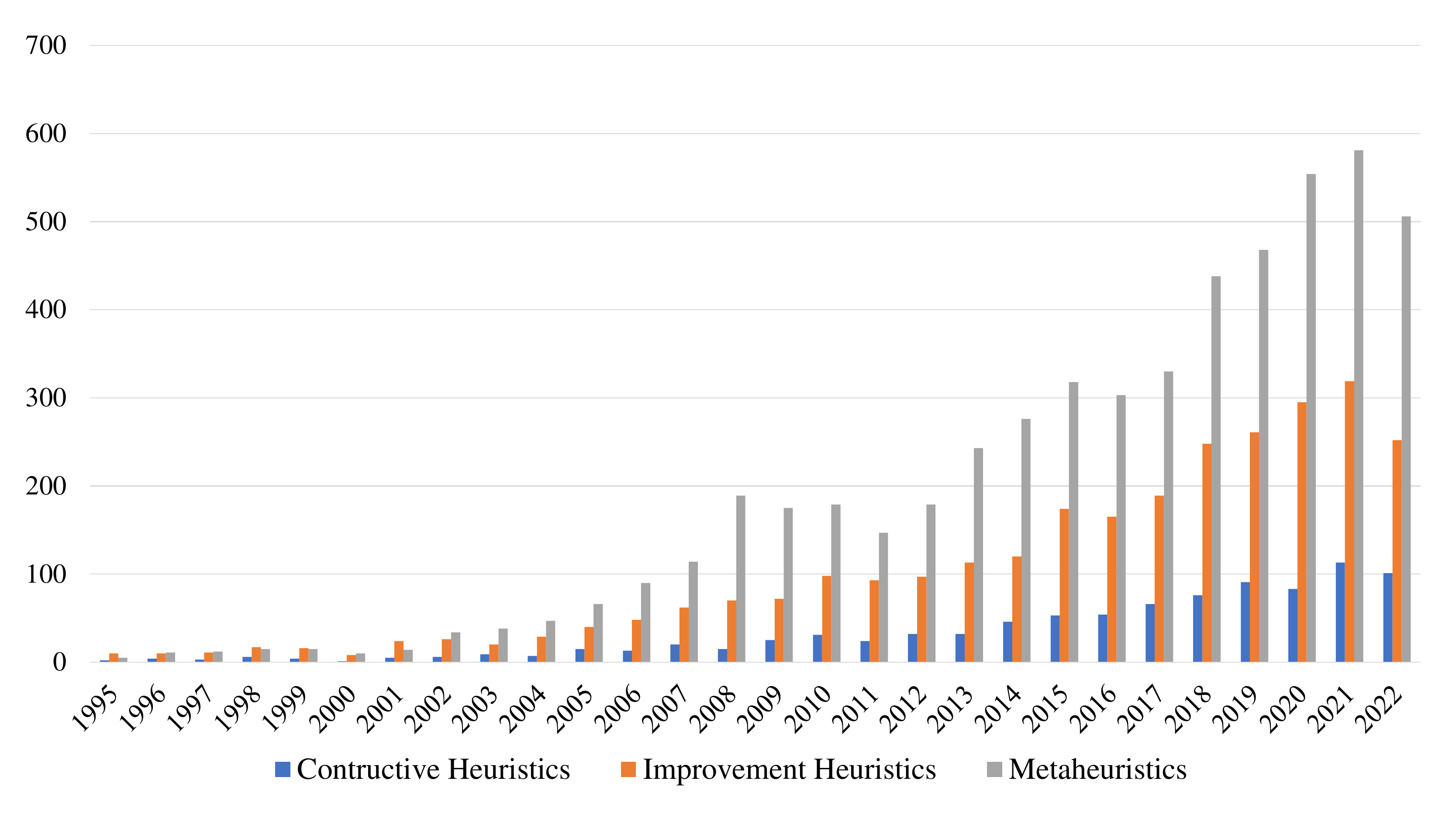}
    \caption{The number of publications on three classes of vehicle routing heuristics.}
    \label{fig:pub_three_categories}
\end{figure}

Fig.~\ref{fig:pub_three_categories} lists the number of publications on three classes of vehicle routing heuristics. It can be observed that the number of publications in all three classes has increased dramatically in the past two decades. There is an urgent need to systematically survey the progress in this active research field and shed light on the insights of their success. This article tends to fill the gap regarding the following aspects:
\begin{enumerate}
    \item We systematically survey the methodologies, recent developments, and applications of vehicle routing heuristics based on the presented classification in Fig.~\ref{fig:heuristic_class}. The shortcomings and strengths of different approaches are discussed.
    \item We review the current state-of-the-art (SOTA) heuristics for some widely studied vehicle routing problems. A general algorithm framework for these methods is presented. We summarize the methods they used for each algorithm component and provide insights into their success.  
    \item We also explore recent progress in developing more general and powerful heuristics. We highlight three emerging research topics: unified heuristic, automatic heuristic design, and machine learning-assisted heuristic. The notable works related to these topics are summarized and some future research directions are discussed.
\end{enumerate}

The rest of this article is organized as follows. Section~\ref{sec2}, Section~\ref{sec3}, and Section~\ref{sec4} provide a review of the methodologies, recent progress, and applications for constructive heuristics, improvement heuristics, and metaheuristics, respectively. In Section~\ref{sec5}, we summarize and discuss the SOTA heuristics for some widely studied vehicle routing problems in a general framework. In Section~\ref{sec6}, we highlight three emerging research topics and review the notable works. Finally, we conclude this paper in Section~\ref{sec7}.

\section{Constructive Heuristics}\label{sec2}
Constructive heuristics construct routing solutions from zero following some fixed empirical procedures. They build a solution quickly, yet there is usually a gap between its result and the optimal one. The existing constructive heuristics can be summarized into four algorithm frameworks: 1) nearest neighbor method, 2) insert method, 3) saving method, and 4) sweep method.

\subsection{Nearest Neighbor Method}	
\begin{figure}[htbp]
    \centering
    \includegraphics[width=0.8\textwidth]{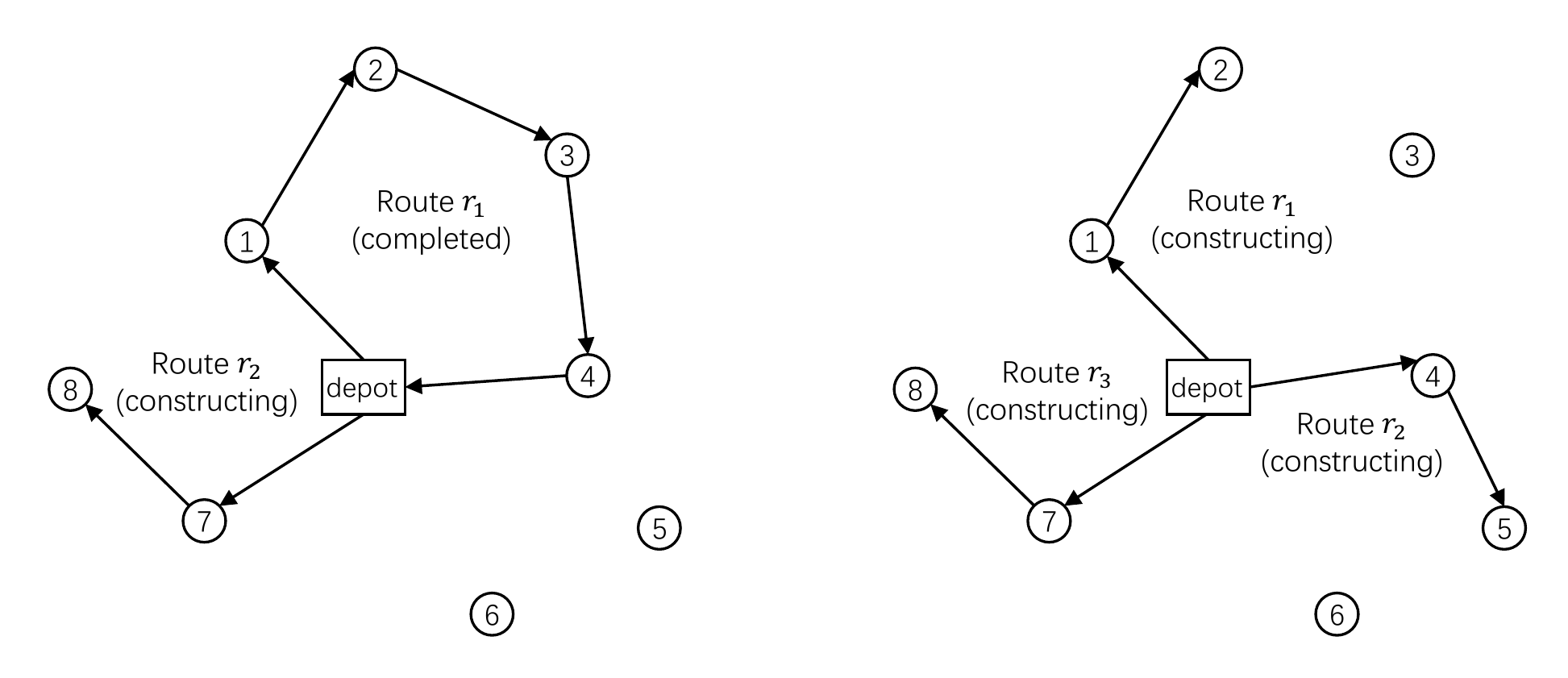}
    \caption{Illustration of sequential (left) and parallel (right) nearest neighbor method.}
    \label{fig:nearest_neighbor}
\end{figure}

The simplest constructive heuristic for VRPs is probably the nearest neighbor method. The routes can be built either sequentially or parallelly. In sequential route building, a route is extended by greedily adding the nearest feasible unrouted customer with the depot as the starting node. A new route is initialized from the depot when no customer can be added. Fig.~\ref{fig:nearest_neighbor} illustrates the nearest neighbor method on a toy example with one depot and eight customers. The first route $r_1$ starts from customer $1$ and ends at customer $4$. Then, a new route $r_2$ is created. The shortcoming of sequential route building is that the last vehicle usually has a lower loading ratio compared to other routes. The shortcoming can be alleviated if the routes are constructed in parallel. In parallel route building, the number of vehicles $K$ is set beforehand and the $K$ routes are extended in parallel. In each iteration, each route is extended with the closest unrouted customer, which means $K$ customers will be added. The process is iteratively performed until all the customers are visited. If customers can not be added to the $K$ routes, then a new route will be created following the sequential route-building strategy. In the toy example in Fig.~\ref{fig:nearest_neighbor}, the number of vehicles is three, and six customers are added to the three routes after two iterations. The time complexity of the nearest neighbor method with $n$ customers is $O(n^2)$, The algorithm iterates $n$ times and takes $O(n)$ to find the nearest neighbor in each iteration.

In recent years, the nearest neighbor method has been integrated into other vehicle routing algorithms, as opposed to being used as an independent heuristic~\cite{joshi2015nearest,mohammed2017solving}. In the tabu search~\cite{du2012combining}, simulated annealing~\cite{vincent2017simulated}, memetic algorithm~\cite{wang2019memetic} and an auxiliary algorithm~\cite{figliozzi2010iterative}, for instance, it serves as the initialization technique. In large neighborhood search~\cite{turkevs2021meta}, it is employed as a recreate operation.

\subsection{Insert Method}	
The basic nearest neighbor method usually has poor performance when more constraints are taken into consideration. For example, if time window constraints are considered, the nearest customer may have the latest time window and the incumbent route can no longer be extended due to the time window constraint. Therefore, the resultant solution will either use an unacceptable number of vehicles or fail to find a feasible solution.

\begin{figure}[htbp]
    \centering
    \includegraphics[width=0.8\textwidth]{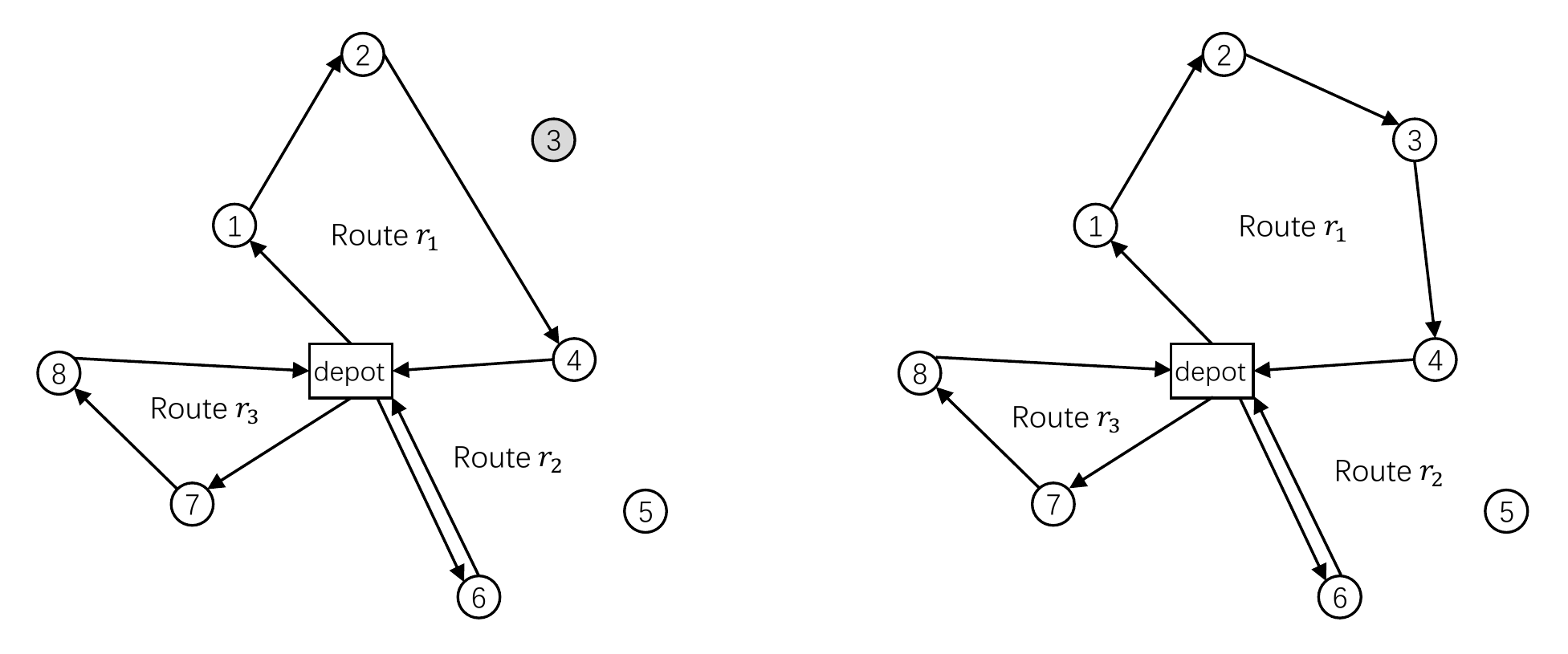}
    \caption{Illustration of one insertion in insert method.}
    \label{fig:insert}
\end{figure}

Insert method overcomes the mentioned situations to some extent. It initializes some empty routes and inserts the unrouted customers one by one into the routes but not necessarily into the end of each route, which is the solution in the nearest neighbor method. In each iteration, the insertion that has the minimum cost is performed. A straightforward implementation of the insert method works as follows. Suppose a route as $R=(v_0,v_1,\dots,v_{n+1})$, where $v_0$ and $v_{n+1}$ denote the depot. The cost of inserting an unrouted customer $v_j$ after customer $v_i,i\in {0,\dots,n}$ is $\Delta_{ij}=c_{ij}+c_{j(i+1)}-c_{i(i+1)}$, where $c_{ij}$ is the cost traveling from customer $i$ to $j$. The cheapest insertion for unrouted customer $v_j$ is defined as $\Delta^\star_j=min\{\Delta_{ij},i\in {0,\dots,n}\}$. The unrouted customer with minimal cost is $j^\star=argmin\{ \Delta^\star_j,j \in U \}$, where $U$ is the set of all unrouted customers. The insert position of $j^\star$ is after $i^\star = argmin\{\Delta_{ij},i\in {0,\dots,n}\}$. Fig.~\ref{fig:insert} illustrates one insert step, where customer $3$ is inserted between customers $2$ and $4$. The worst time complexity of the insert method is $O(n^3)$. Similar to the nearest neighbor method, the insert method iterates $n$ times. However, there are two layers of minimization problem in each iteration: first, to find the minimal insertion cost for each unrouted customer and then find the unrouted customer with the cheapest minimal insertion cost.

The mentioned insert method can be served as the greedy version. Farthest insert and regret insert are two main revisions of the greedy one. Farthest insert~\cite{rosenkrantz1977analysis,renaud2000heuristic} inserts the farthest unassigned customer into the best position in the incumbent routes in each iteration. The regret insert~\cite{shaw1998using,pisinger2007general} extends the greedy insert. In each iteration, instead of performing the current best insertion, regret insert chooses to insert the node with the largest regret. The regret is defined as the difference between the cost of inserting the node in the best position and the $i$th-best position, where $i$ is usually set to be a small number. The idea is that the insertion with large regret will lead to a high cost if it is not inserted in the best position. Applications of this adjustment can be found in many works, including~\cite{diana2004new,wang2020iterative,dumez2021large}.

Several hybrid implementations are carried out to improve the straightforward insert methods. For example,~\cite{christofides1979vehicle} first applied a sequential route construction procedure to estimate the number of routes and then used a parallel procedure to generate a better solution. \cite{balseiro2011ant} used insert heuristics to assist the ant colony algorithm for time-dependent VRP. \cite{archetti2016vehicle} used it to construct an initial solution. The insert methods were also used implicitly as the recreate operations in large neighborhood search~\cite{pisinger2007general}.

\subsection{Saving Method}	
\begin{figure}[htbp]
    \centering
    \includegraphics[width=0.8\textwidth]{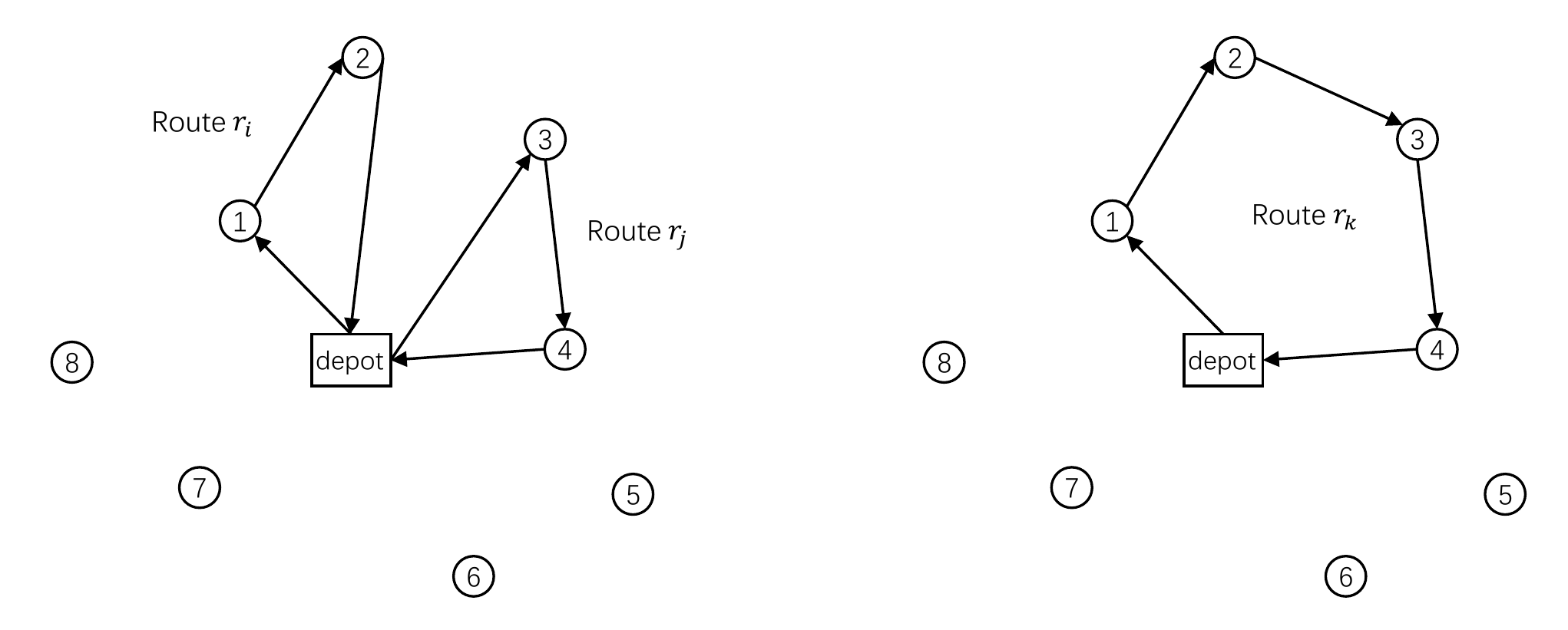}
    \caption{Illustration of one merger in saving method.}
    \label{fig:saving}
\end{figure}

Saving method, proposed in 1964~\cite{clarke1964scheduling}, might be the best-known constructive heuristic. It starts with an initial solution, in which a different route serves each customer, and iteratively combines the short routes into longer routes with lower overall costs. The procedure can be performed in parallel or in sequence. In its parallel version, the method iteratively combines two routes with the endpoints $i$  and  $j$, which produces the maximum feasible distance saving  $s_{ij}=c_{i0}+c_{0j}-c_{ij}$. In the sequential version, each route is considered in turn. One route is iteratively extended by feasible saving operation until no such feasible merge can be used. Fig.~\ref{fig:saving} shows one merger, where route $r_i$ and route $r_j$ are merged to form route $r_k$. Note that four possible routes can be generated by different combinations of route $r_i$ and route $r_j$ with respect to different directions.

The time complexity of a straightforward implementation of the saving method is $O(n^3)$. The algorithm starts with $n$ routes. It takes $n$ merge operations (iterations) and each merge operation is decided at the cost of $O(n^2)$ if every maximum feasible distance saving $s_{ij}=c_{i0}+c_{0j}-c_{ij}$ is checked or rechecked in each iteration. To reduce the complexity, the $n(n-1)/2$ combinations corresponding to $n(n-1)/2$ possible mergers can be computed and sorted at the beginning of the algorithm. The sort can be finished in $O(n^2log_2 n)$ using a fast sorting algorithm.

Among early revisions of the saving method,~\cite{gaskell1967bases} and~\cite{yellow1970computational} parametrized the original equation of saving and proposed a general equation  $s_{ij}=c_{i0}+c_{0j}-\lambda c_{ij}$, where $\lambda$ is the parameter to adjust the importance of the distance between two endpoints compared with the distance from endpoints to depot. \cite{segerstedt2014simple} only considered the first-pair calculated savings and used this information to iteratively construct routes. In addition, the saving method can be combined with other concepts. For example, an insert method based on the saving method was proposed by~\cite{mole1976sequential}, which not only combines two routes but also allows inserting nodes into a route. Matching-based saving algorithms~\cite{desrochers1989matching,altinkemer1991parallel,wark1994repeated} served the choosing of route pairs as a matching problem, where the match weight can be the saving of distances or other criteria. They usually produce better results than the straightforward saving method with higher computational costs. 

More recently, the saving method has been extended and fine-tuned by other methods. \cite{reimann2004d} designed a divide-and-conquer procedure with a saving-based ant system for solving vehicle routing problems. \cite{battarra2008tuning} used a genetic algorithm to optimize the basic saving method. \cite{juan2010sr} proposed a stochastic version of the classic savings heuristic. Instead of creating new routes by selecting the largest relevant savings from the savings list, it selects the savings using a geometric distribution and a probabilistic function. \cite{juan2011use} used Monte Carlo simulation, cache, and splitting techniques to improve the saving heuristics. Moreover, instead of only considering the distance in the selecting of saving, \cite{caccetta2013improved} and~\cite{cengiz2022fuzzy} take into account multiple criteria.

Because of its simplicity and flexibility in implementation, the saving method has been extended to solve 
dynamic VRP~\cite{dror1986stochastic}, multi-depot VRP with time windows~\cite{tamashiro2010tabu},VRP with simultaneously pickup and delivery~\cite{gajpal2010saving}, stochastic VRP with time windows~\cite{wang2016three}, real-world problems~\cite{grasas2013vehicle}, and humanitarian aid distribution~\cite{cengiz2022fuzzy}.

\subsection{Sweep Method}	
\begin{figure}[htbp]
    \centering
    \includegraphics[width=0.8\textwidth]{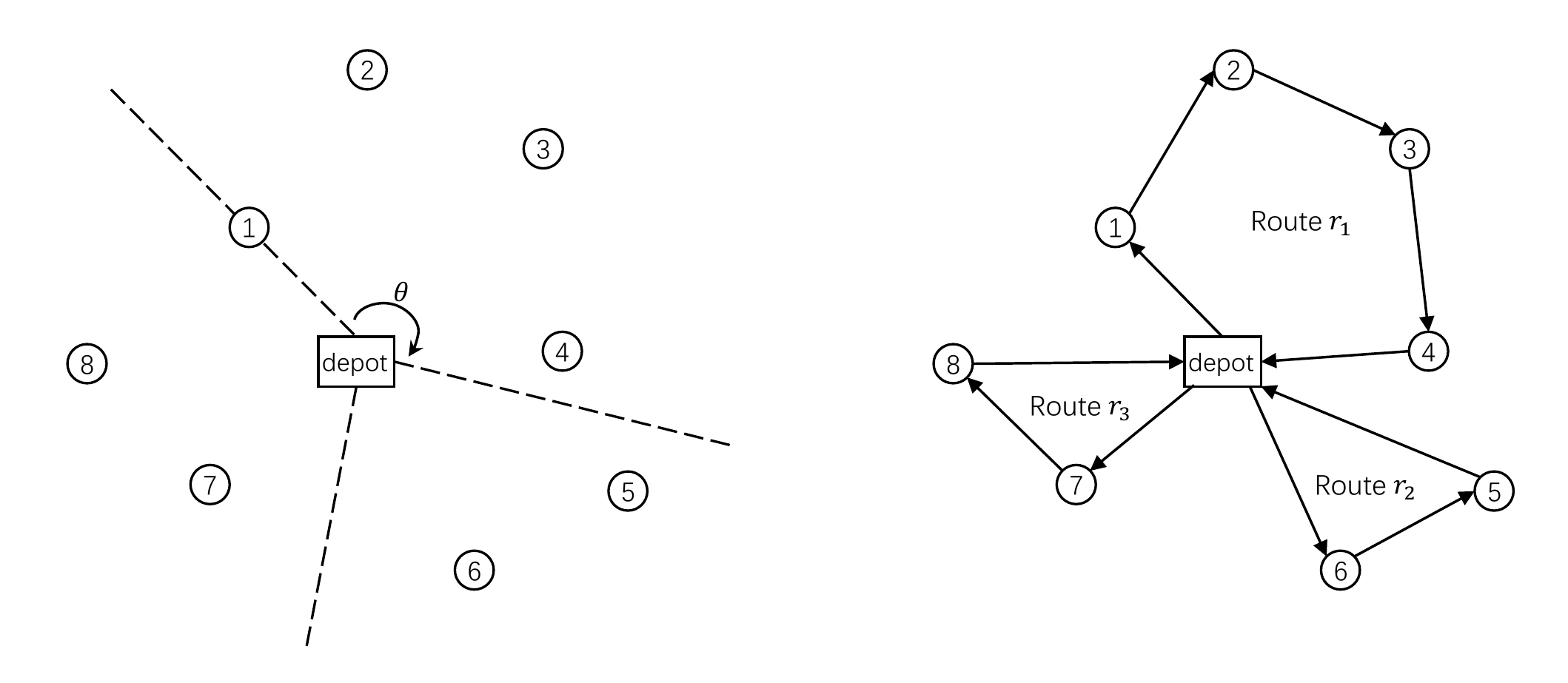}
    \caption{Illustration of sweep method.}
    \label{fig:sweep}
\end{figure}

Some early works that used the concept of sweep method can be found in~\cite{lampkin1972computers,wren1972computer} while the sweep algorithm is commonly attributed to the seminal work in~\cite{gillett1974heuristic}. The algorithm first sets the depot as the origin and sorts the nodes according to the polar angle. In its basic version, the node is added to the route circularly. A new route will be created if the insertion is infeasible. Another approach is a cluster-first, route-second method. All the customers are clustered into several clusters according to the polar angle and a TSP is solved in each cluster, as illustrated in Fig.~\ref{fig:sweep}. After a feasible solution is created, some improvement heuristics such as $\lambda$-opt in~\cite{gillett1974heuristic} can be used.  The performance of the sweep method can be greatly affected by the location of the depot. A poor result will be generated when the depot is off-centered. It can be tackled by using various reference points~\cite{na2011some} rather than only the depot. In addition, the last route often has a lower loading ratio because the clusters are built one by one. The time complexity of the basic sweep method is $O(n)$ because all the required computing is adding the next unrouted customer to the previous route according to the polar angle. In more advanced versions, the time complexity mainly depends on the method used for the TSPs.

Among others~\cite{shin2011centroid,peya2018capacitated}, sweep nearest algorithm~\cite{na2011some} and distance-based sweep nearest algorithm~\cite{zahrul2019distance} are two straightforward implementations of the basic sweep method. In each cluster, they select the next nearest customer instead of  until reach the capacity. The difference is the former starts from the unassigned customer with the smallest polar angle while the latter starts from the farthest unassigned customer. Both algorithms outperform the basic sweep algorithm and the results are competitive with modern heuristics.

Sweep method has been extended to solve various VRP variants. \cite{zhishuo2005sweep,akhand2017capacitated,thammano2021hybrid} applied it on capacitated VRP. \cite{renaud2002sweep} proposed a petal-based algorithm, where the customer set of petals is determined based on the sweep, for mix vehicle routing problem. \cite{suthikarnnarunai2008sweep} used a sweep heuristic method with local search and an integer
programming model for split delivery VRP. \cite{dondo2013sweep} integrated a sweep heuristic into the optimization method to solve VRP with cross-docking. \cite{euchi2021hybrid} combined sweep and genetic algorithms to solve the VRP with drones.

\section{Improvement Heuristics}\label{sec3}

Improvement heuristics explore the neighborhood of the incumbent solution to achieve an objective improvement. They can quickly converge to the local optimal and thus are efficient at solving large-scale routing problems. There are two widely recognized classes of improvement heuristics: intra-route and inter-route. The difference between the two classes is the neighborhood structure. The former searches inside a single route, while the latter involves multiple routes.

\subsection{Intra-route Method}

Intra-route improvement heuristics explore the neighborhood involving only one route. Most of them originated from the local search operators for TSP~\cite{applegate2011traveling}. For instance, among the simplest ones, one customer can be relocated to a different position in the same route, or two customers in one route can be exchanged. The $\lambda$-opt heuristic, as a more general operation on edges, removes $\lambda$ edges from a route and recreates $\lambda$ other edges to connect the disjoint sequences. As a full implementation of $\lambda$-opt moves requires  $O(n^\lambda)$ for $n$ customers, a small $\lambda$ is often used, such as 2-opt, 3-opt, and Or-exchange.

\begin{figure}[htbp]
    \centering
    \subfigure[Relocate]{                 \includegraphics[width=0.25\textwidth]{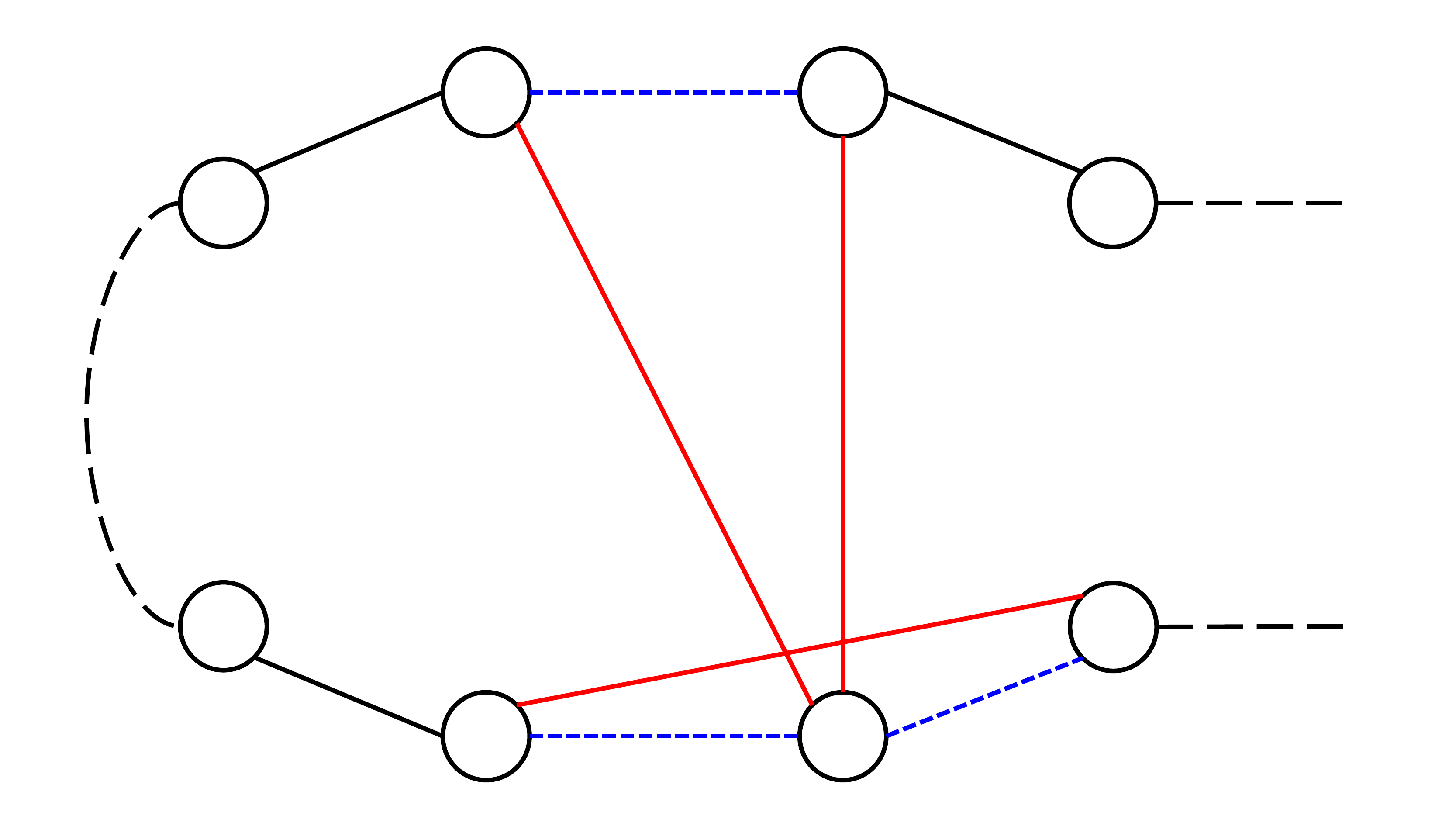}
    }
    \subfigure[Exchange]{                 \includegraphics[width=0.25\textwidth]{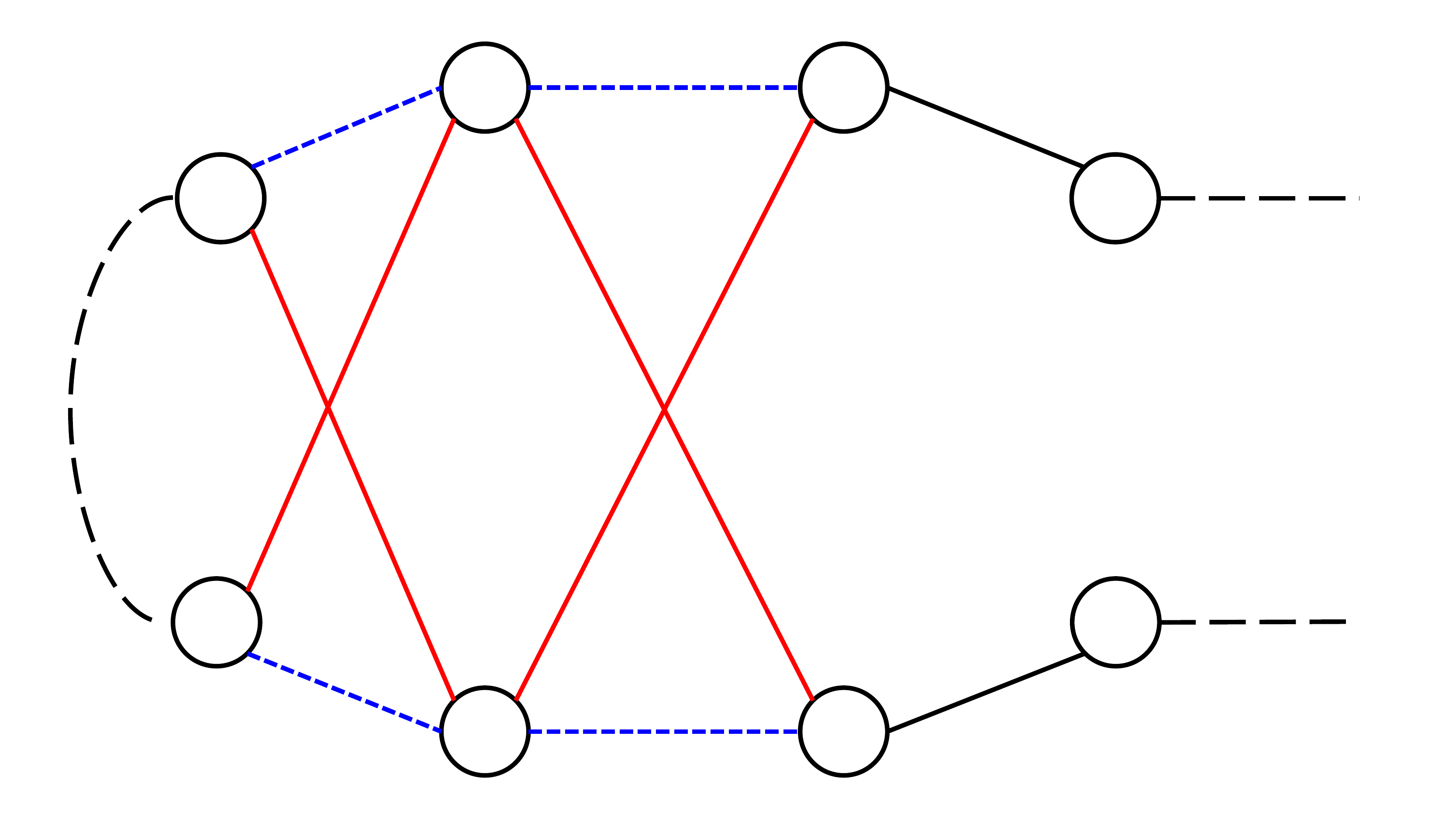}
    }
    \subfigure[2-opt]{                 \includegraphics[width=0.25\textwidth]{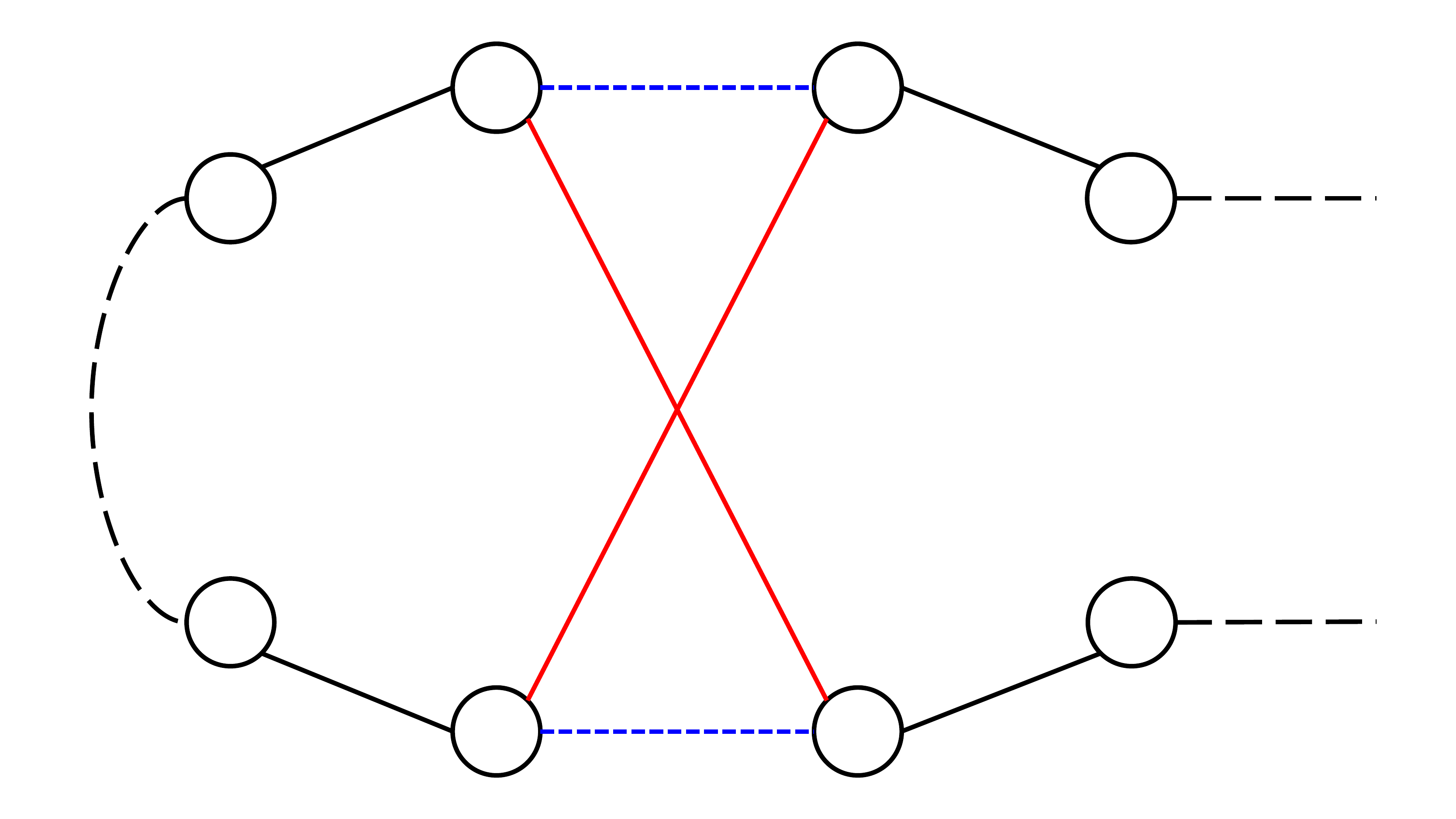}
    }
    \subfigure[3-opt]{                 \includegraphics[width=0.25\textwidth]{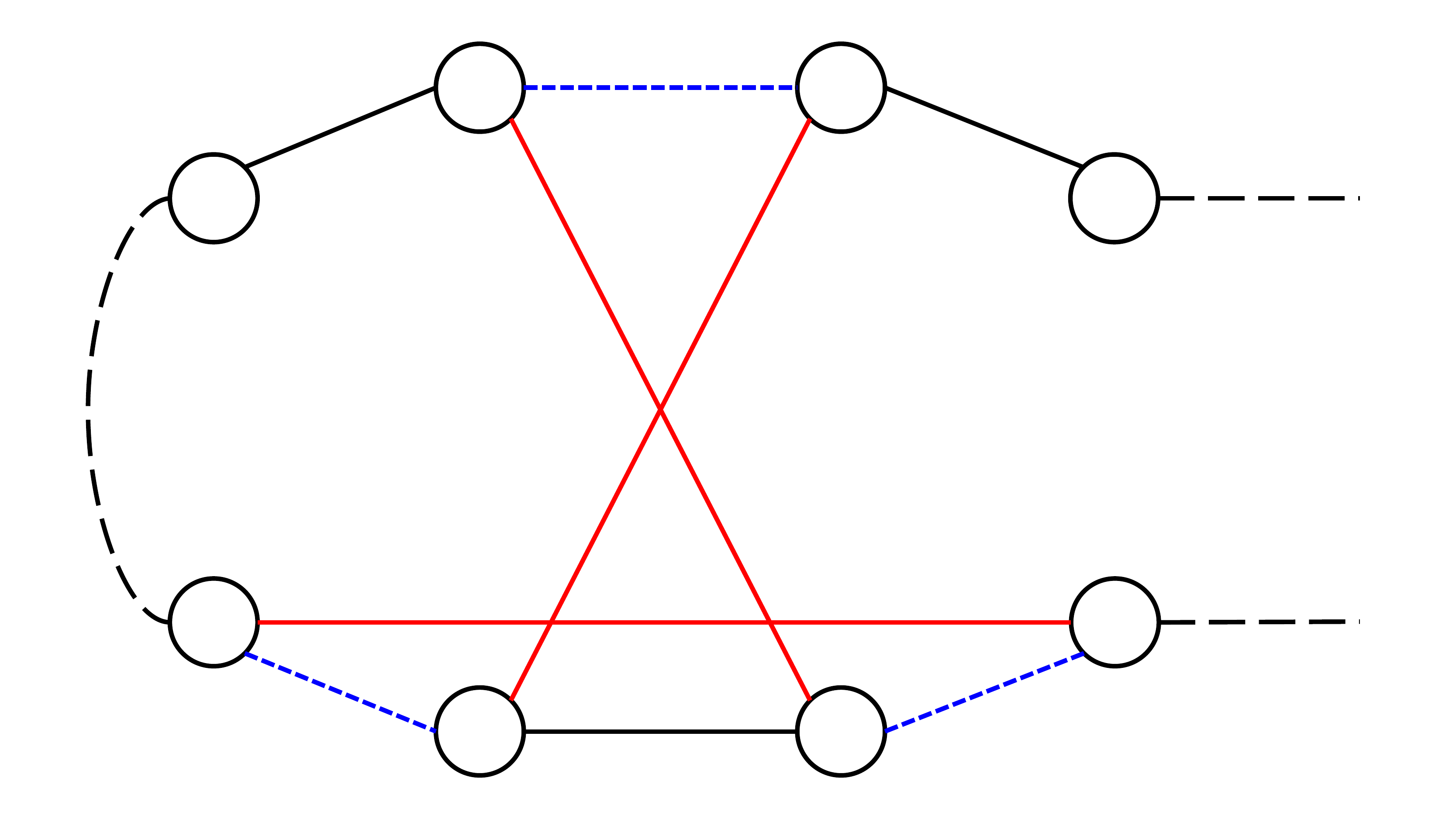}
    }
    \subfigure[OR-exchange]{                 \includegraphics[width=0.25\textwidth]{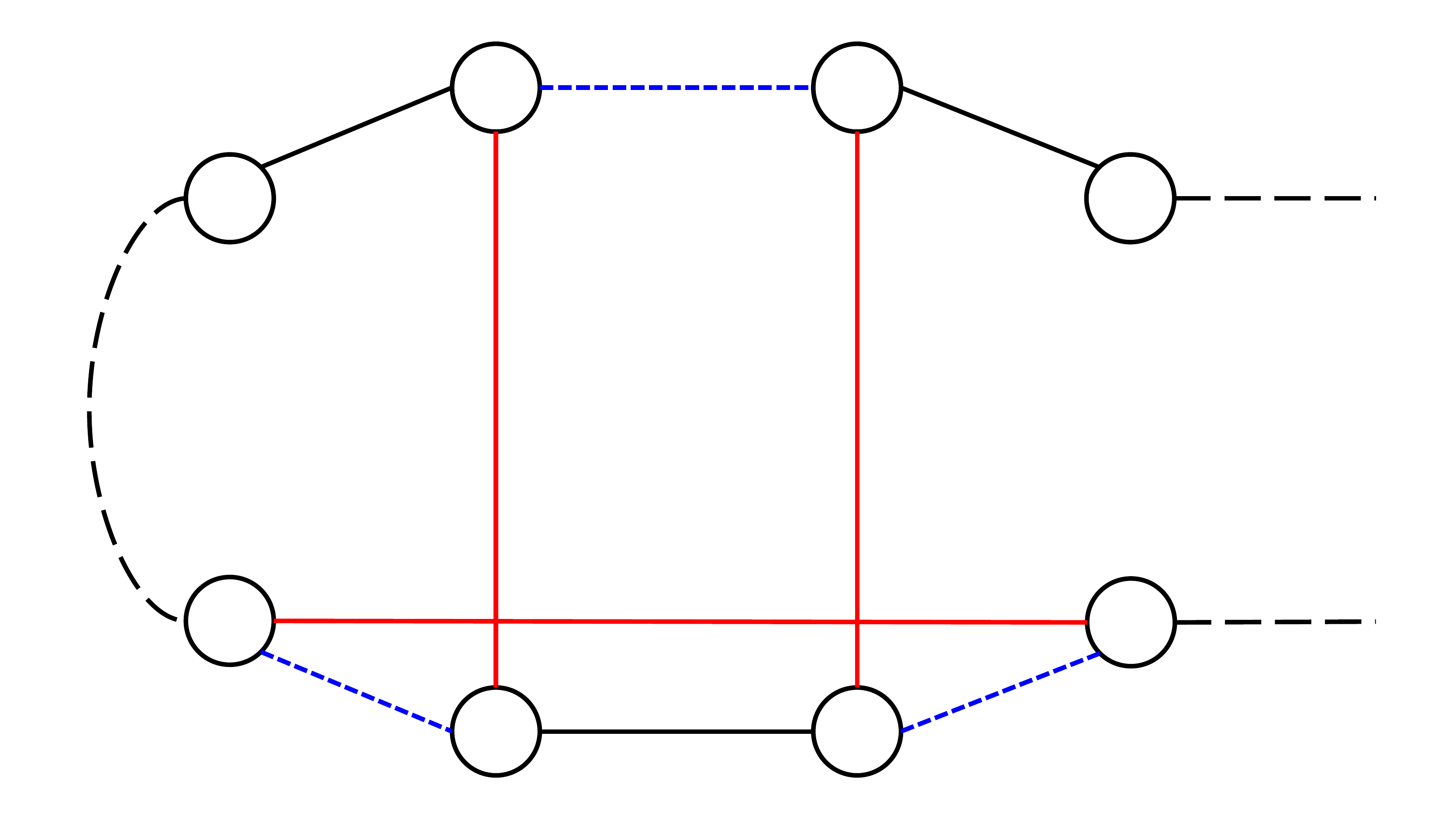}
    }
    \subfigure[GENI]{                 \includegraphics[width=0.25\textwidth]{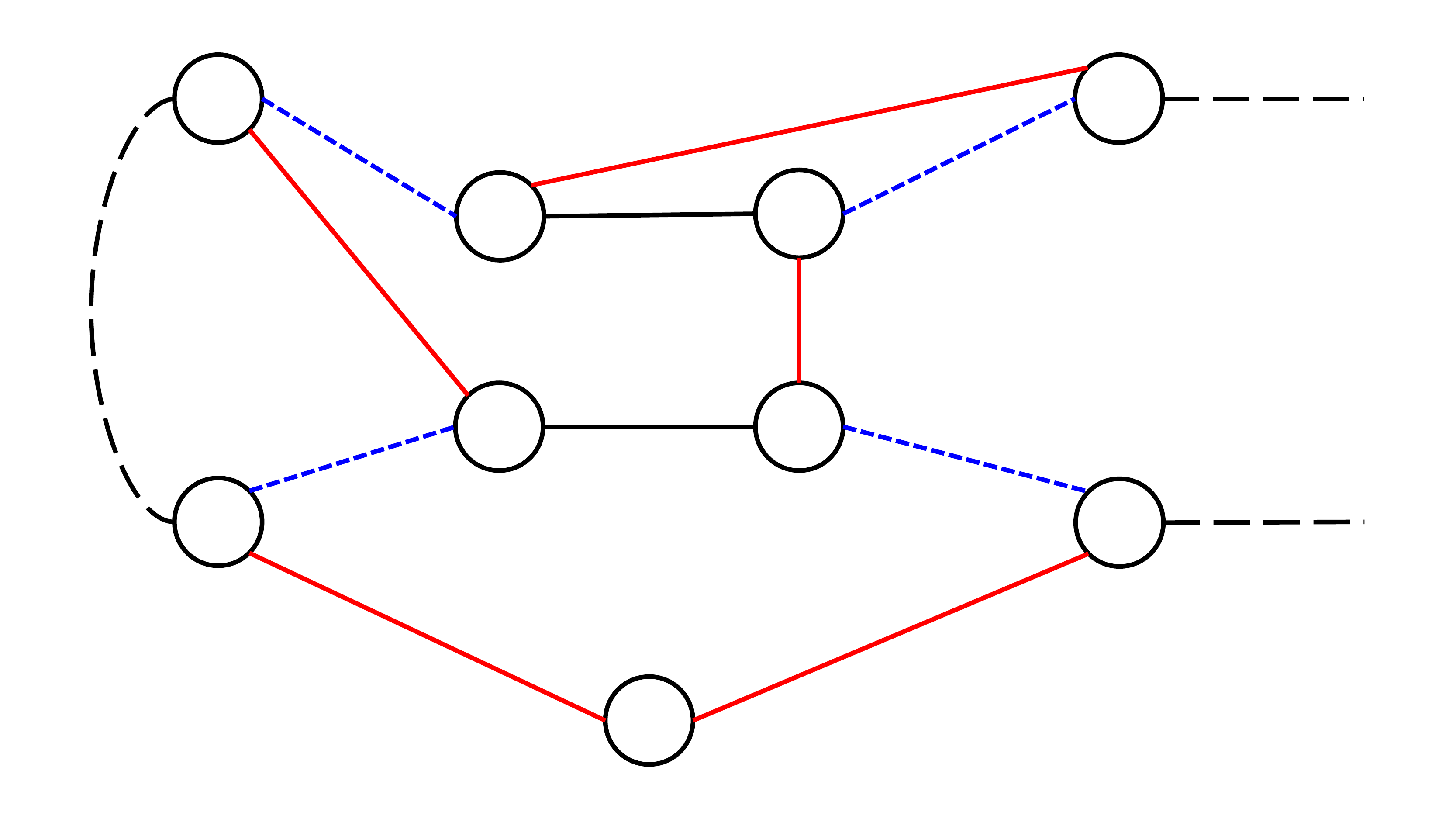}
    }
    \caption{Illustration of different intra-route improvement heuristics.}
    \label{fig:IH_intra-route}

\end{figure}

Fig.~\ref{fig:IH_intra-route} illustrates the intra-route improvement heuristics. The nodes are customers and the lines are routes. The dotted blue lines are segments of the routes removed by the improvement heuristics and the red lines are the segments added by the improvement heuristics. The descriptions for these heuristics are given as follows:

\begin{itemize}
    \item  \textit{Relocate:} Relocate selects one route from the solution and relocates one node to another position in the route. The time complexity of finding the best relocation solution is $O(n^2)$.
    \item  \textit{Exchange:} Exchange selects one route and exchanges the positions of two nodes in the route. The time complexity of finding the best relocation solution is $O(n^2)$.
    \item \textit{2-opt,3-opt,$\lambda$-opt:} 2-opt first selects one route and then reconnects the ends of two edges in the route~\cite{lin1965computer}. In other words, it reverses a sequence of nodes within the route. The time complexity of finding the best reconnection is $O(n^2)$. The 2-opt can be regarded as a subset of 3-opt, which reconnects three edges with a time complexity of $O(n^3)$. $\lambda$-opt further generalizes them to considering $\lambda$ edges~\cite{lin1965computer}.
    \item  \textit{Or-exchange:} Or-exchange selects a sequence of nodes with a preset length and relocates it to another position in the same route~\cite{or1976traveling}. The or-exchange can be regarded as a subset of 3-opt. Different from 3-opt, it only needs to specify the location of the start (or end) node of the sequence and the corresponding destination in the route. Therefore, the time complexity of finding the best exchange is $O(n^2)$.
    \item  \textit{GENI:} GENI selects one route and a subset of three vertices. For a vertex $v$ not yet on the route, it implements the least cost insertion considering the two possible orientations of the tour and the two insert types~\cite{gendreau1992new}. The time complexity of each insertion is $O(np^4+n^2)$ when the neighborhood size is $p$.
\end{itemize}

\subsection{Inter-route Method}

\begin{figure}[htp]
    \centering
    \subfigure[Insert]{                 \includegraphics[width=0.28\textwidth]{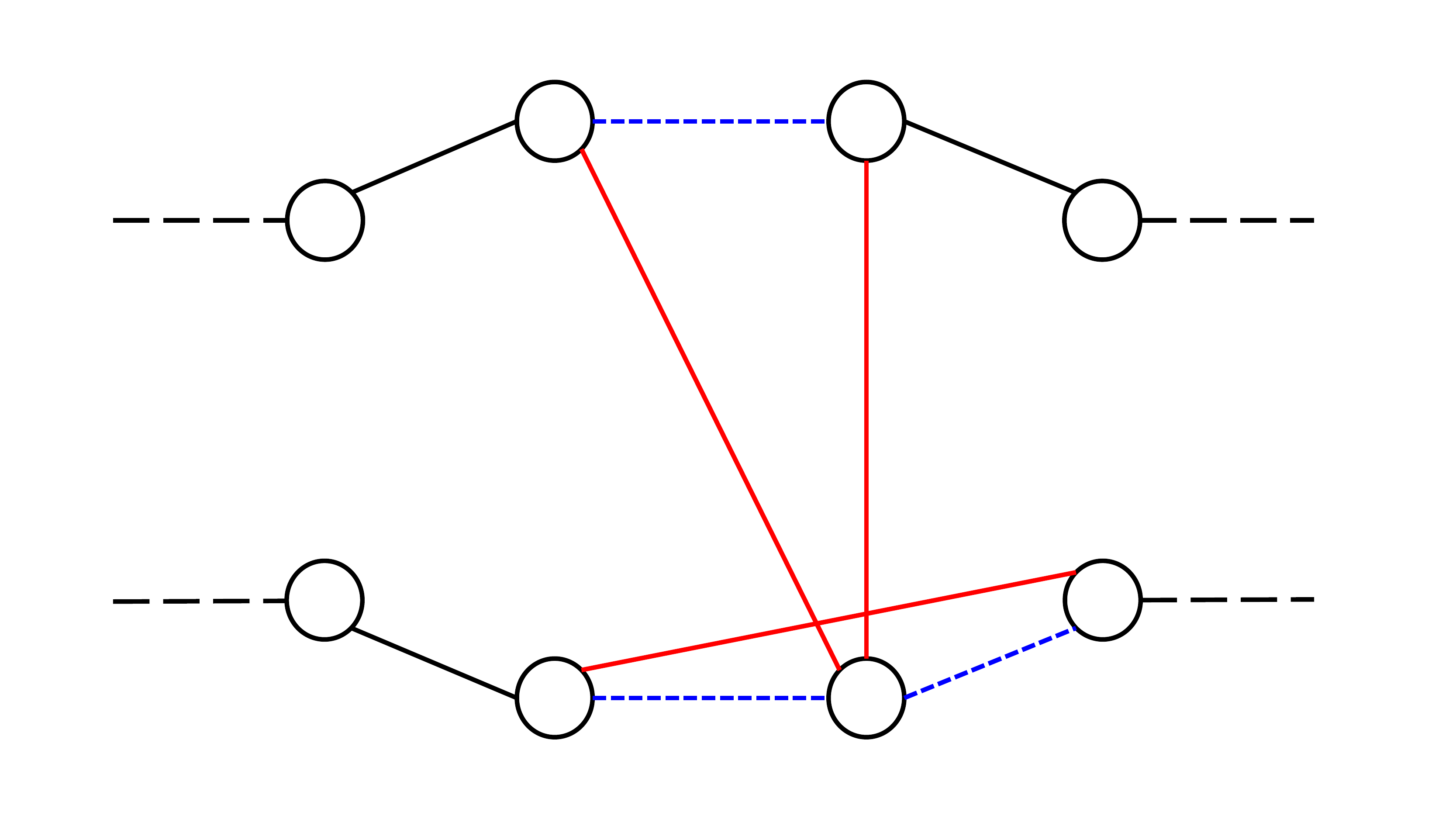}
    }
    \subfigure[Swap]{                 \includegraphics[width=0.28\textwidth]{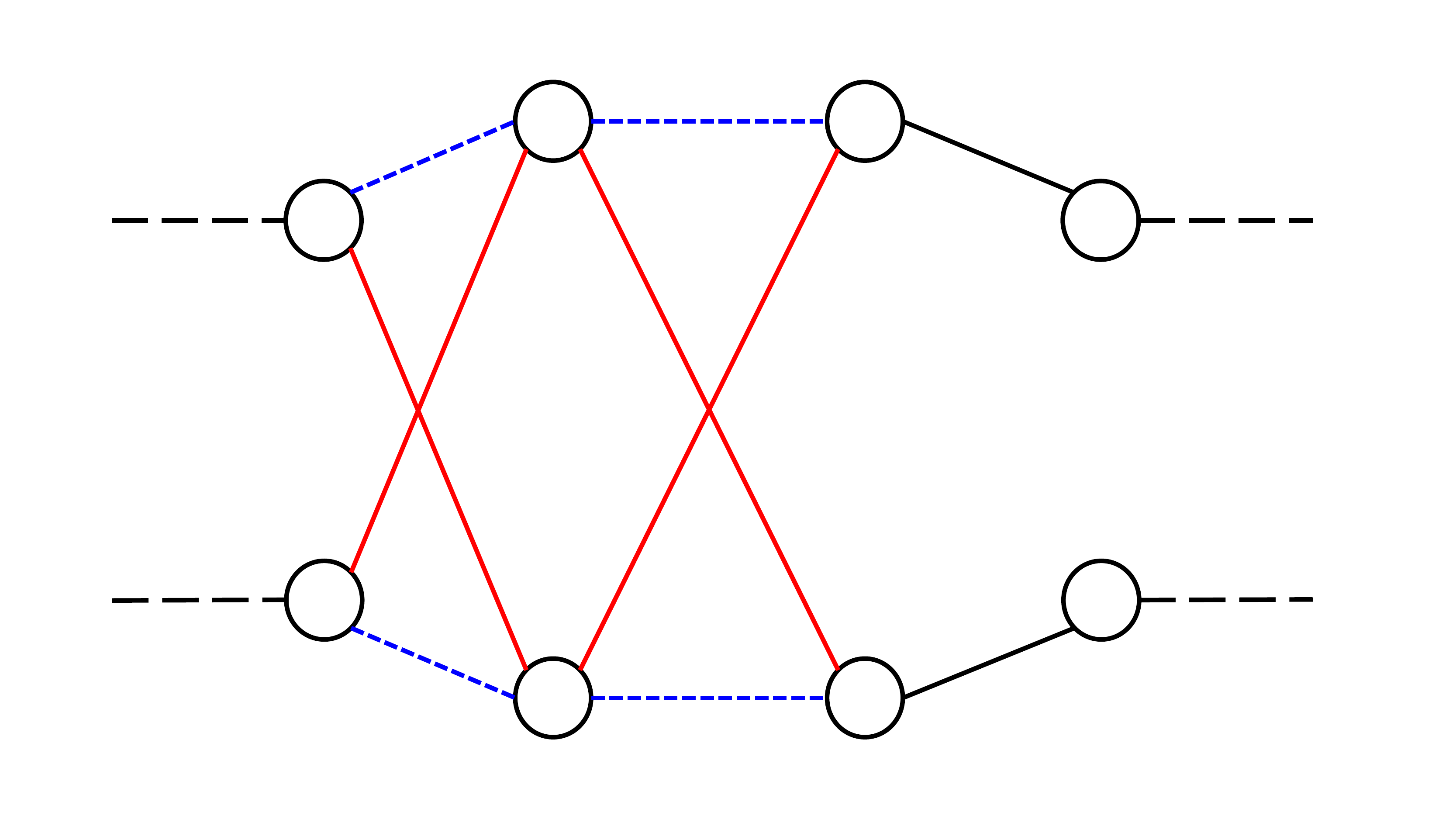}
    }
    \subfigure[2-opt*]{                 \includegraphics[width=0.28\textwidth]{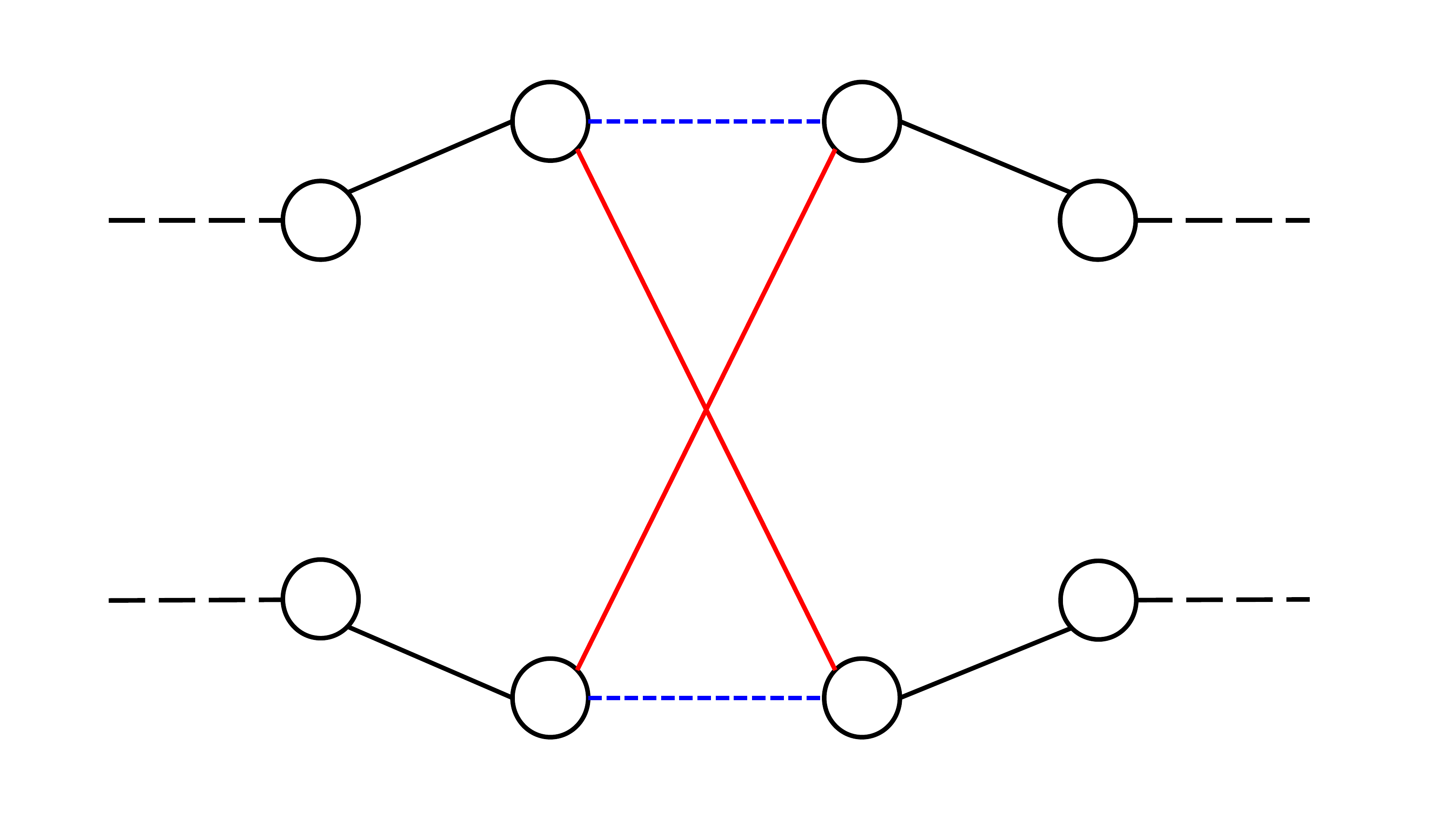}
    }
    \subfigure[CROSS]{                 \includegraphics[width=0.3\textwidth]{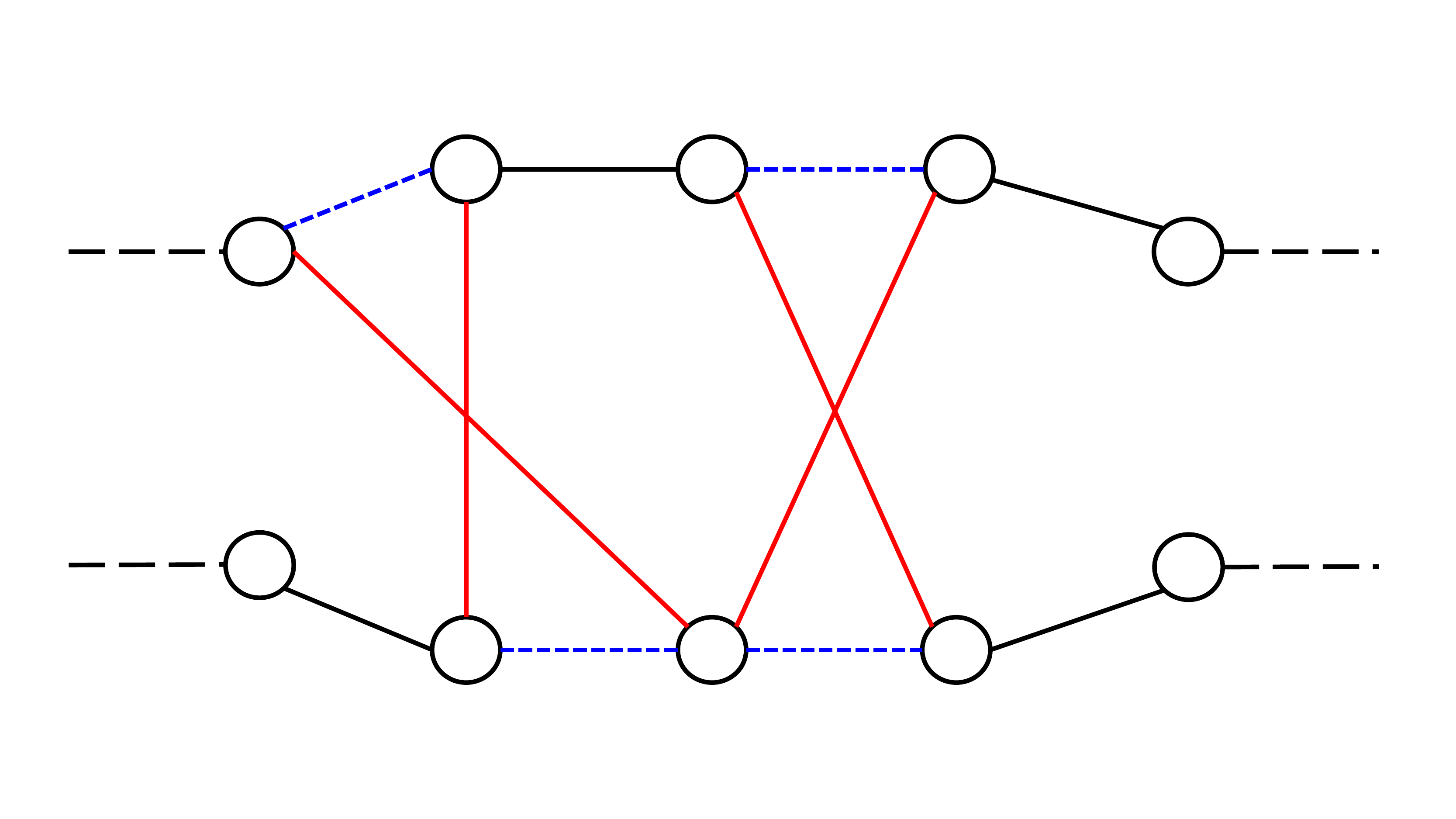}
    }
    \subfigure[$\lambda$-interchange]{                 \includegraphics[width=0.3\textwidth]{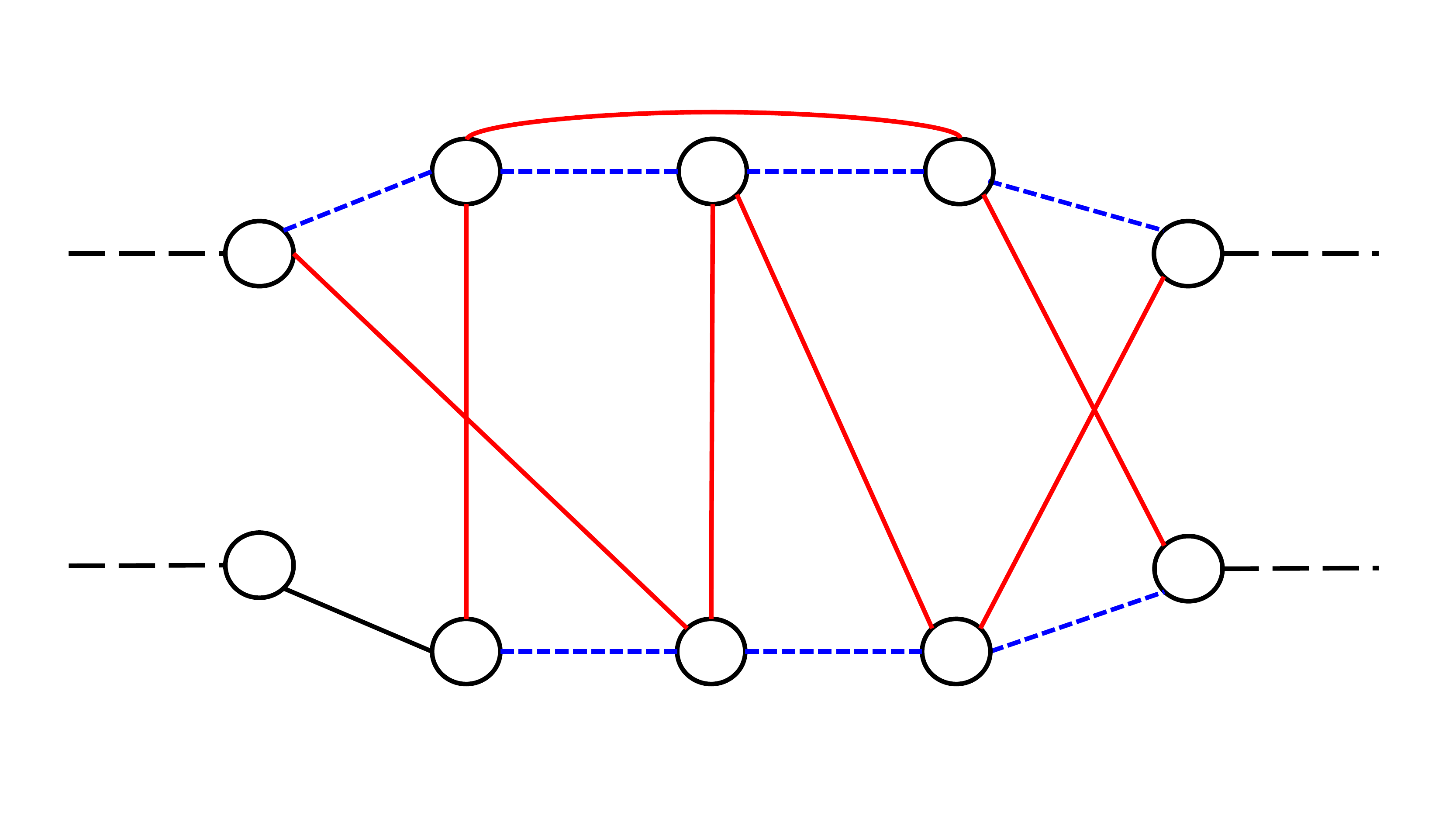}
    }
    \caption{Illustration of different inter-route improvement heuristics.}
    \label{fig:IH_inter-route}
\end{figure}

Inter-route heuristics involve local searches across multiple routes. Many of them are extensions of intra-route counterparts. For example, insert and swap are the extension of relocate and exchange, respectively. The former removes a customer from one route and reinserts it into another route. The latter swaps two customers from different routes. In addition, the exchange of two edges from different routes is denoted as 2-opt* to distinguish it from the well-known 2-opt. CROSS exchanges two strings, each of them having at most $\lambda$ customers. $\lambda$-interchange further generalizes CROSS. It allows exchanging any set of less than $\lambda$ nodes between two routes, even if they are non-consecutive. Fig.~\ref{fig:IH_inter-route} illustrates some representative inter-route improvement heuristics, which are introduced as follows:

\begin{itemize}
    \item  \textit{2-opt$^{\star}$:} 2-opt selects two routes and reconnects the end of two edges~\cite{potvin1995exchange}. The reconnection with the best cost reduction is chosen. If the cost of the selection of two routes can be neglected, its time complexity is the same as 2-opt.
    \item  \textit{Insert:} Insert selects one node and inserts it into another position~\cite{osman1993metastrategy}. Similar to the Relocation operator, the time complexity is $O(n^2)$.
    \item  \textit{Swap:} Swap selects two nodes and swaps their locations~\cite{osman1993metastrategy}. Like the Exchange operator, the time complexity is $O(n^2)$. \cite{vidal2022hybrid} designed an extended version of the Swap, named Swap*. It consists in exchanging two customers between different routes without an insertion in place. Swap* contains more improving moves than Swap but its direct implementation requires a time complexity of $O(n^4)$. \cite{vidal2022hybrid} proposed an algorithm to reduce the complexity to sub-quadratic.
    \item  \textit{CROSS:} CROSS exchanges two customer sequences (one of the two sequences can be empty)~\cite{taillard1997tabu}. It generalizes the three mentioned operators: 2-opt$^{\star}$, Insert, and Swap. The time complexity is at most $O(n^4)$ and $O(\lambda^2n^2)$ when the size of the sequence is limited by a value $\lambda$. $\lambda$ = 3 is often used to keep an acceptable computational cost.
    \item  \textit{$\lambda$-interchange:} $\lambda$-interchange further generalizes CROSS~\cite{osman1993metastrategy}. It allows exchanging any set of less than $\lambda$ nodes between two routes. The set can be non-consecutive, empty, and reversed during the reinsert.
\end{itemize}

\begin{figure}[htbp]
    \centering
    \subfigure[Intra-route]{                 \includegraphics[width=0.4\textwidth]{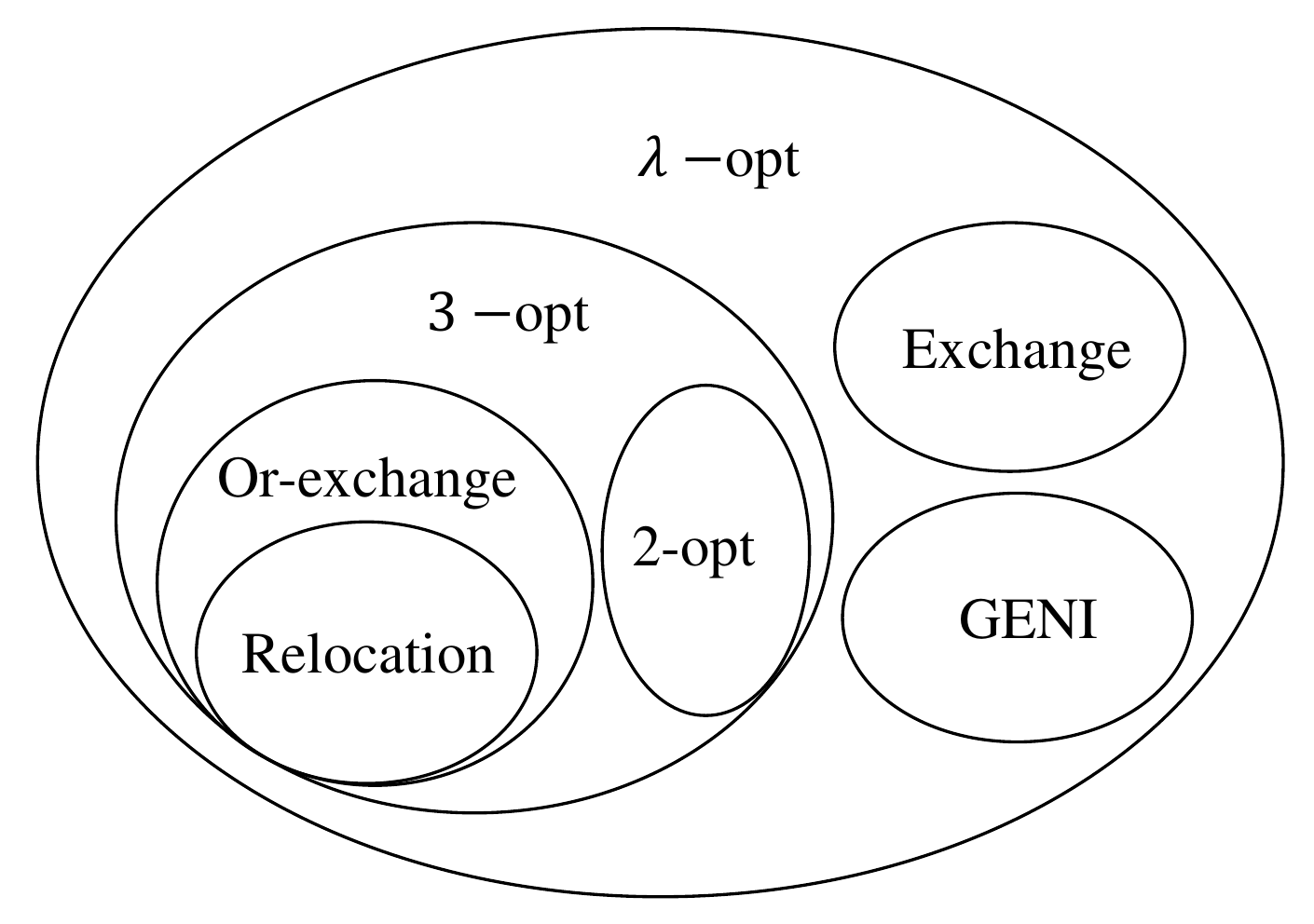}
    }
    \subfigure[Inter-route]{                 \includegraphics[width=0.4\textwidth]{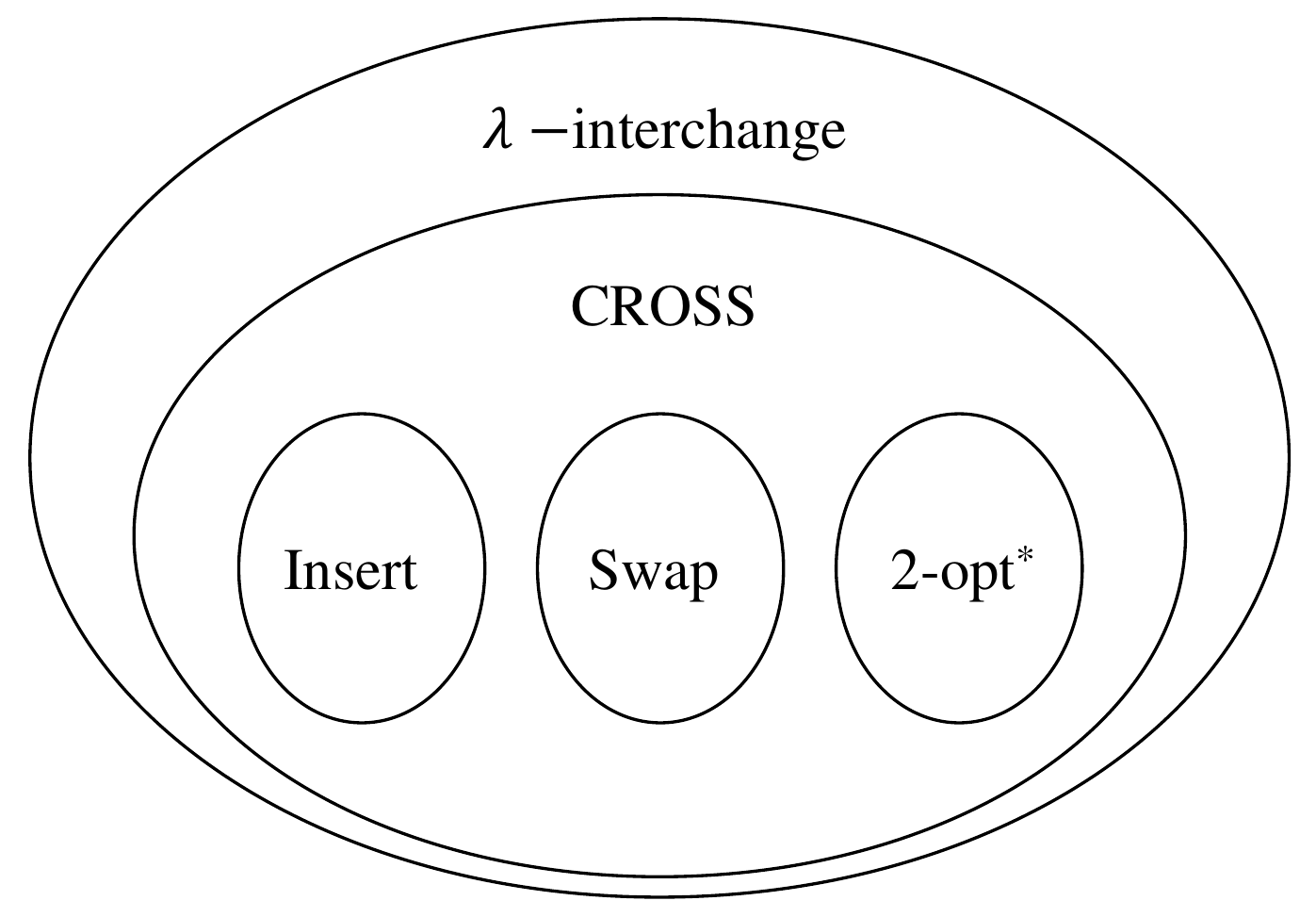}
    }
    \caption{Relationships between improvement heuristics.}
    \label{fig:IH_relationships}
\end{figure}

Fig.~\ref{fig:IH_relationships} illustrates the relationships between improvement heuristics. The bigger circle represents a larger searching neighborhood and also worse time complexity. Among intra-route methods, $\lambda$-opt is the most general one. It includes all the other heuristics as its subsets. For inter-route methods, all the heuristics can be regarded as special cases of $\lambda$-interchange. These heuristics are essentially different trade-offs between the computational cost and the solution quality. A common belief is that a local optimum for one local search heuristic typically will not be a local optimum for another~\cite{arnold2019knowledge}. In recent years, the research works in this regard have primarily focused on using different improvement heuristics together~\cite{vidal2014unified,penna2019hybrid,accorsi2021fast}, rather than selecting any one of them.

Improvement heuristics have made some interesting steps forward in recent years. One approach is to use machine learning to improve algorithm performance. For example,~\cite{d2020learning} used policy deep learning to learn a 2-opt improvement heuristic. \cite{kim2022neuro} leveraged graph neural network to predict the cost-decrements in CROSS heuristic and utilized the predicted cost-decrements as guidance for search to reduce the search cost. Instead of learning to improve one improvement heuristic,~\cite{wu2021learning} used reinforcement learning to select dynamically from three different improvement heuristics and the results demonstrated how well it outperforms the conventional approach. Another approach is to explore the neighborhood efficiently. Swap* enlarged the neighborhood of Swap and integrated multiple improving moves. The efficiency of Swap* was improved by a new updating algorithm~\cite{vidal2022hybrid}.

\section{Metaheuristics}\label{sec4}

Metaheuristics are presented in a more general and high-level way~\cite{sorensen2013metaheuristics}, in contrast to constructive heuristics and improvement heuristics, which are problem-dependent and attempt to exploit the feature and structure of the target problem. They are widely recognized as an effective approach for many hard optimization problems~\cite{gendreau2010handbook,weinand2022research}. Due to their efficiency and scalability, metaheuristics are becoming increasingly dominating in vehicle routing research~\cite{elshaer2020taxonomic}.

According to the population management strategy, we classify metaheuristics into two categories: 1) single-solution-based and 2) population-based methods, as illustrated in Fig.~\ref{fig:heuristic_class}. Single-solution-based (also named as neighborhood-based~\cite{vidal2012hybrid} or local search-based~\cite{funke2005local}) methods iteratively perform low-level searches on one incumbent solution. The low-level search is typically improvement heuristics. Except for the design or selection of low-level heuristics, the key will be how to guide the search process in a principled way to dump out of the local optimum and retain a certain level of exploration of the search space. It is achieved by tailored high-level rules. There are three approaches for designing the high-level rules: 1) multiple starts, 2) changing the landscape, and 3) designing the acceptance rule. In the first approach, the local search starts from multiple initial points. The strategies include searching from multiple starts, changing start points during optimization, and combining both. In the second approach, local search is guided by changing the fitness landscape. They either modify the landscape during optimization or use different landscapes. In the last approach, a well-designed acceptance criterion is used to drive the search forward. They leave out some visited unpromising solutions or accept worse solutions with a certain probability.

Evolutionary algorithms and swarm intelligence are two essential approaches for doing population-based searches. Evolutionary algorithm, such as genetic algorithm~\cite{gendreau2010handbook}, is inspired by biological evolution. Individuals who are stronger (more suited to the environment) than the competition are more likely to generate offsprings who can better survive. A hot research area in recent years is applying multiobjective evolutionary algorithms (MOEA) to solving VRPs with multiple objectives~\cite{wang2015multiobjective,wang2018two,li2018decomposition,zajac2021objectives,zarouk2022novel,qi2022qmoea}. Swarm intelligence~\cite{kennedy2006swarm} is a field of study that examines natural and artificial systems made up of numerous individuals that cooperate with each other. The decentralized, collective, and self-organized cooperative behavior of social entities like bee colonies~\cite{karaboga2014comprehensive}, ant colonies~\cite{dorigo2006ant}, fish, and others is the foundation for cooperation. Rather than using a single metaheuristic concept, the third category integrates various kinds of heuristics. Memetic algorithm~\cite{neri2011handbook} is one typical example, which combines population-based methods with other efficient search heuristics.

\begin{figure}[htbp]
    \centering
    \includegraphics[width=0.8\textwidth]{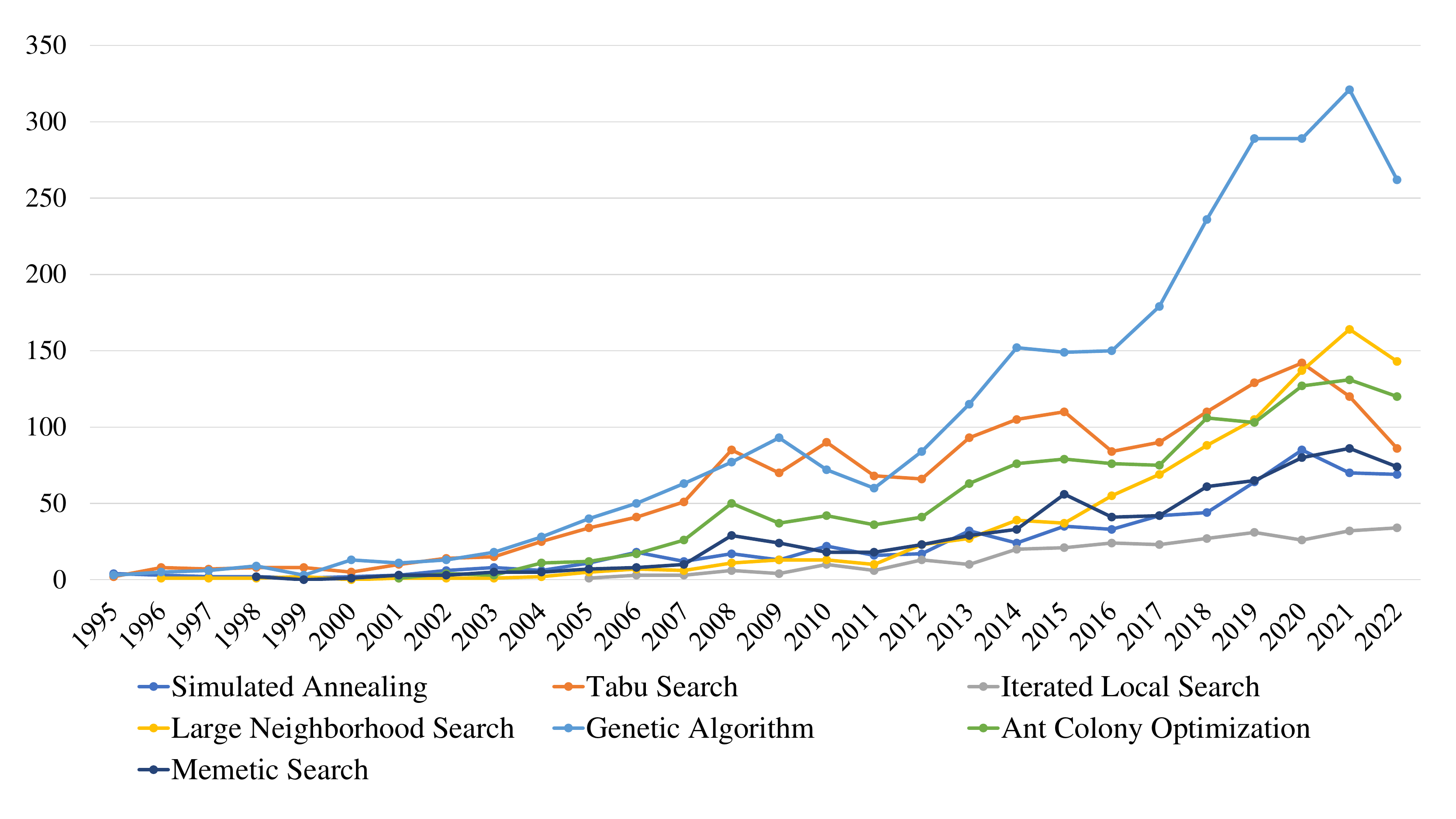}
    \caption{The number of publications of different metaheuristics in vehicle routing research.}
    \label{fig:number_publication_metaheuristics}
\end{figure}

In this section, we survey the widely-used metaheuristics in vehicle routing, including simulated annealing (SA), tabu search (TS), iterated local search (ILS), large neighborhood search (LNS), genetic algorithm (GA), ant colony optimization (ACO), memetic algorithm (MA). These algorithms are among the top-used algorithms in vehicle routing as reviewed in~\cite{elshaer2020taxonomic} and provide good coverage of different sub-classes of metaheuristics in Fig.~\ref{fig:heuristic_class}. To keep this article concise, we do not include other metaheuristics, such as variable neighborhood search (VNS)~\cite{polacek2004variable,kytojoki2007efficient,hemmelmayr2009variable,khouadjia2012comparative,cai2022variable}, guided local search (GLS)~\cite{kilby1999guided,arnold2019efficiently}, greedy randomized adaptive search process (GRASP)~\cite{kontoravdis1995grasp,prins2009grasp,haddadene2016grasp,bruglieri2022grasp}, and particle swarm optimization (PSO)~\cite{ai2009particle,gong2011optimizing,goksal2013hybrid,marinakis2019multi,islam2021hybrid}, to mention a few. Fig.~\ref{fig:number_publication_metaheuristics} shows the number of publications of the selected metaheuristics in vehicle routing research, which are collected from the Web of Science. Among them, GA received the most studies, followed by TS, LNS, and ACO. The number of publications on LNS and MA is growing rapidly in the last decade. We present a brief introduction to their methodologies and review their recent updates and applications.
 
\subsection{Single-Solution-Based Method}

\subsubsection{Simulated Annealing}

Simulated annealing was first introduced by researchers in the early 1980s for combinatorial optimization~\cite{kirkpatrick1983optimization}. The idea is similar to the actual annealing of materials in physics. In the SA algorithm, a mechanism is used to mutate the incumbent solution to generate a new solution in the design space. The mechanism is commonly one or multiple local search operators. The solution itself and the fitness of the solution correspond to the state and the energy in the annealing process, respectively. A typical strategy used in SA to accept a new solution is as follows: 1) a new solution is accepted if its fitness is better than the old one, 2) if the new solution is worse than the old one, it is accepted with a probability of $exp(\frac{f(s_{new})-f(s_{old})}{T})$, where the control parameter $T$ represents the temperature. The process is iteratively performed with a decreasing temperature. The key of SA is on the acceptance criterion, it escapes the local optimal by accepting the slightly worse solution and converges to the optimal solution with the decreasing temperature.

The first application of SA on vehicle routing was carried out in~\cite{osman1993metastrategy}, which proposed a hybridization method with tabu search for solving capacitated VRP. \cite{van1995improvement} proposed an enhanced heuristic based on the SA to solve VRP. \cite{chiang1996simulated} applied the SA for solving a VRP with time-window constraints, where two different neighborhood structures were investigated and the annealing process was enhanced with a short-term memory function via tabu list. After that, a number of works have been carried out to apply SA to various VRP variants, such as VRP with an independent route length~\cite{tavakkoli2006hybrid}, truck and trailer routing problem~\cite{lin2009solving}, VRP with two-dimensional loading constraints~\cite{leung2010simulated}, VRP with time windows and synchronization constraints~\cite{afifi2013simulated}, open location-routing problem~\cite{vincent2015simulated}, and hybrid VRP~\cite{vincent2017simulated}. More recently, 
\cite{vincent2021adaptive} proposed a new variant of simulated annealing with an adaptive mechanism for selecting neighborhood moves. \cite{ilhan2020population,ilhan2021improved} designed an improved population-based simulated annealing algorithm for capacitated VRP. A population of solutions is maintained during optimization. Each solution is improved using simulated annealing heuristics and the solutions incorporate each other through crossover operators.

\subsubsection{Tabu Search} 

Tabu search~\cite{glover1986future} was proposed in 1986 to provide a principled strategy to avoid local search algorithms trapped in local optimal and guide the search towards a promising direction. An important concept is the tabu lists, which keep track of the most recent history of the search and prevent the search from going back to previously visited solutions. 

The early attempts of applying tabu search on vehicle routing include~\cite{osman1993metastrategy,gendreau1994tabu,gendreau1999parallel,toth2003granular}. Several improvements to the basic tabu search framework have been applied to solving vehicle routing problems. For example, some works~\cite{zachariadis2010adaptive,repoussis2010solving,li2010adaptive} used adaptive memory programming in tabu search. The adaptive memory programming exploits a collection of strategic memory components and is more efficient than using a single fixed memory management procedure~\cite{glover1997tabu}. \cite{toth2003granular} proposed granular tabu search. The granular is a restricted neighborhood, which not containing the moves that are has a low probability of leading to a good feasible solution. For vehicle routing problems, the granular is typically defined as a limited number of "short" arcs, based on the observation that "long" arcs are less possible to be part of high-quality solutions. Because of the simple yet effective idea of granular tabu search, it has been widely applied on VRPs in recent years~\cite{prins2007solving,kirchler2013granular,goeke2019granular} and the idea of granular has been used in other local search-based methods~\cite{branchini2009adaptive}.

As a general concept, tabu search can be easily integrated into other algorithms~\cite{thangiah1994hybrid,flisberg2009hybrid,belhaiza2014hybrid}. Recently, \cite{wang2017two} combined tabu search with simulated annealing in the local search to efficiently solve a VRP with cross docks and split deliveries. \cite{schermer2019hybrid} proposed a hybrid method based on a variable neighborhood search framework and the tabu list is adopted in a local search procedure to avoid cycling. \cite{sadati2021efficient} used tabu search in the shaking procedure of variable neighborhood search to solve a class of multi-depot VRPs.

The applications of tabu search include multi-depot VRP~\cite{renaud1996tabu}, periodic VRP~\cite{cordeau1997tabu}, VRP with heterogeneous fleet~\cite{gendreau1999tabu}, and open VRP~\cite{fu2005new}, heterogeneous multi-type fleet VRP~\cite{wang2015heuristic}, multi-depot open VRP~\cite{soto2017multiple}, and VRP with discrete split deliveries and pickups~\cite{qiu2018tabu}, among others~\cite{gmira2021tabu,sadati2021efficient}.

\subsubsection{Iterated Local Search}

Iterated local search builds on the simple idea that iteratively generates a sequence of solutions using the underlying heuristics~\cite{baum1986iterated,johnson1997traveling,applegate2003chained}. Different from SA and TS, ILS uses perturbation to jump out of local optimal instead of changing fitness or acceptance criterion. The perturbation can be as simple as a random restart or more commonly a principled strategy, such as various neighborhood search heuristics.

The applications of ILS can be found on the prize-collecting VRP~\cite{tang2006iterated}, time-dependent VRP~\cite{hashimoto2008iterated}, VRP with time penalty functions~\cite{ibaraki2008iterated}, VRP with heterogeneous~\cite{penna2013iterated}, VRP with backhauls~\cite{cuervo2014iterated}, VRP with multiple, incompatible commodities and multiple trips~\cite{cattaruzza2014iterated}, split delivery VRP~\cite{silva2015iterated}, location-routing~\cite{nguyen2012multi}, and weighted VRP~\cite{wang2020iterative}. Several recent improvements are carried out with additional procedures to extend the basic ILS. For example, \cite{brandao2020memory} proposed a memory-based ILS for multi-depot open VRP. The history of optimization is used for the definition of the perturbation procedures. \cite{sabar2021population} designed a population-based ILS, where a population with promising solutions is maintained. It adaptively performs a local search algorithm with various evolutionary operators to solve dynamic VRP. \cite{maximo2021hybrid} and \cite{maximo2022ails} developed an adaptive ILS with a diversity control method for capacitated VRP. The results are quite competitive, especially on large-scale problems.

\subsubsection{Large Neighborhood Search}

Large neighborhood search, proposed by~\cite{shaw1997new,shaw1998using}, is developed based on the idea that a large neighborhood has a large probability containing high-quality local optimum. Although any neighborhood can be adopted in the LNS paradigm, for vehicle routing problems, it typically consists of two main procedures: ruin and recreate (i.e., destroy and repair). Ruin removes a part of the incumbent solution and recreate reinserts the removed part into the partial solution to form a new solution. The ruin method typically takes into consideration the relatedness of removed customers. The relatedness is usually measured by distance and other similarities. The insert procedure usually uses constructive heuristics, such as insert methods.

Adaptive Large Neighborhood Search (ALNS)~\cite{ropke2006adaptive,ropke2006unified,pisinger2007general} is a well-known extension of LNS. ALNS consists of multiple ruin and recreate operators and adaptively selects these operators in each iteration. The probability of choosing each operator is determined using the historical performance during the optimization. Generally, the adaptation scheme promotes performance. While~\cite{christiaens2020slack} recently designed an LNS heuristic without multiple operators and the weight adaptation. It only used an adjacent string removal and a greedy insert with blinks for ruin and recreate, respectively, and outperformed other cutting-edge methods. Results showed that simplicity and reproducibility are not always necessarily come at the expense of solution quality. LNS as well as ALNS has been successfully applied to many vehicle routing problems~\cite{pisinger2019large,hemmelmayr2012adaptive,grangier2017matheuristic,grangier2016adaptive,chen2021adaptive,friedrich2022adaptive}.

\subsection{Population-Based Method}

\subsubsection{Genetic Algorithm} 

Genetic algorithm has been a popular method in the optimization community for many years~\cite{holland1975adaptation,de1975analysis,gendreau2010handbook}. It builds on the idea that natural evolution drives the evolution of all species and maintains a good balance of population diversity and adaptiveness. The two key evolutionary operations are crossover and mutation. The mutation involves permutation on one solution to generate new offspring. In a broader sense, any permutation operators can be regarded as a kind of mutation. The crossover is an inter-solution exchange, which typically generates offsprings from two selected parents. Most of the crossover operators used in vehicle routing inherit from the ones used in genetic algorithms for TSP~\cite{larranaga1999genetic}. Some frequently used operators in vehicle routing are 1) Order Crossover~\cite{davis1985applying}, 2) Partially Mapped Crossover~\cite{goldberg1985alleles}, 3) Edge Recombination Crossover~\cite{whitley1989scheduling}, 4) Cycle Crossover~\cite{oliver1987study}, and 5) Alternating Edges Crossover~\cite{grefenstette1985genetic}, to name a few. A comparison study of different crossover operators can be found in~\cite{puljic2013comparison,varun2017study}.

The research of using GA for solving vehicle routing took off twenty years ago~\cite{hwang2002improved,baker2003genetic,berger2003hybrid}. Along with the application study of GA on different modern VRP variants, such as multi-depot VRP~\cite{lau2009application}, pickup and delivery problem~\cite{tasan2012genetic}, green VRP~\cite{da2018genetic} and multiobjective VRPTW~\cite{ombuki2006multi}, new crossover and mutation operators are proposed. \cite{vidal2012hybrid} designed a periodic insertion crossover to accept solutions that violate vehicle capacity and take both solution quality and diversity into account. \cite{kool2022hybrid} proposed a selective route exchange, which exchanges two sets of randomly selected similar routes from two parents. Different from other methods, it preserves depot visits and the split procedure~\cite{vidal2016split} is saved. Because the classical genetic search framework is found to be not enough aggressive on combinatorial optimization problems, combining GA with different search techniques is a popular trend that helps traditional GA perform better on challenging routing problems. The integrated methods include particle swarm~\cite{marinakis2010hybrid,kuo2012hybrid}, simulated annealing~\cite{ariyani2018hybrid}, and sweep~\cite{euchi2021hybrid}.

\subsubsection{Ant Colony Optimization}

Ant colony optimization~\cite{dorigo1999ant,dorigo2006ant,dorigo1999ant,gendreau2010handbook} is motivated by the behavior of real ants. They communicated with each other using pheromones, which are updated during the search procedure. Similar to the strategy used by the ant colony, ACO constructs solutions to the problem at hand based on the pheromones, which they then adjusted during the optimization to take into account their search history. The first example of such an algorithm is ant system~\cite{dorigo1992optimization,dorigo1991ant,dorigo1991positive,dorigo1996ant} and its early applications on TSP~\cite{dorigo1999ant}. 

The seminal work of~\cite{bell2004ant} extended ACO to vehicle routing problems to allow the search of multiple routes. \cite{yu2009improved} applied an improved ACO, which adopted an ant-weight strategy to update the increased pheromone. Recently, many works are carried out to combine ACO with other algorithms to promote the performance on hard vehicle routing problems~(e.g., \cite{zhang2019hybrid,kyriakakis2021hybrid,stodola2020hybrid}). The applications include multi-compartment VRP~\cite{reed2014ant}, VRP with simultaneous pickup and delivery~\cite{kalayci2016ant}, heterogeneous VRP with mixed backhaul~\cite{wu2016label}, multi-depot green VRP with multiple objectives~\cite{li2019improved}, multiobjective VRP with flexible time windows~\cite{zhang2019hybrid},  periodic VRP with a time window and service choice~\cite{wang2020improved}, dynamic VRPs~\cite{mavrovouniotis2015ant,xiang2021pairwise}, and capacitated electric VRP~\cite{jia2021bilevel}.

\subsubsection{Memetic Algorithm}

The memetic algorithm has shown promise in a number of application areas, particularly efficient for hard optimization problems including vehicle routing problems. Unlike other classical population-based methods, MA tasks the advantages of different searching approaches. It is stimulated by the process of cultural evolution, where information is not simply exchanged between individuals but also processed and enhanced during communication. This enhancement is commonly achieved by incorporating efficient search methods, such as local search techniques~\cite{norman1991competitive,neri2011handbook}.

The implementations of MA on vehicle routing typically combine population-based heuristics with improvement heuristics. The cooperation of different search approaches usually results in an efficient algorithm. Actually, many cutting-edge vehicle routing algorithms are the implementation of MA or use the concept of MA~\cite{tang2009memetic,mei2011memetic,zhang2017memetic,decerle2019hybrid,cattaruzza2014memetic,el2008memetic,nagata2010penalty}. Among them, hybrid genetic search (HGS)~\cite{vidal2012hybrid,vidal2013hybrid}, which combines genetic search and local search operators and utilizes a diversity control on the population, is a representative one. It is regarded as the most competitive vehicle routing heuristic with its recent revisions~\cite{penna2019hybrid} and enhancements~\cite{vidal2022hybrid,santana2022neural,kool2022hybrid}. Recent applications of MAs include heterogeneous VRP with time windows~\cite{molina2020acs}, generalized bike-sharing rebalancing problem~\cite{lu2020effective}, open VRP under uncertain travel times~\cite{sun2021light}, and unmanned aerial vehicle routing~\cite{xiang2021effective}.

\section {Insights into SOTA Heuristics}\label{sec5}

\begin{figure}[htbp]
    \centering
    \includegraphics[width=0.5\textwidth]{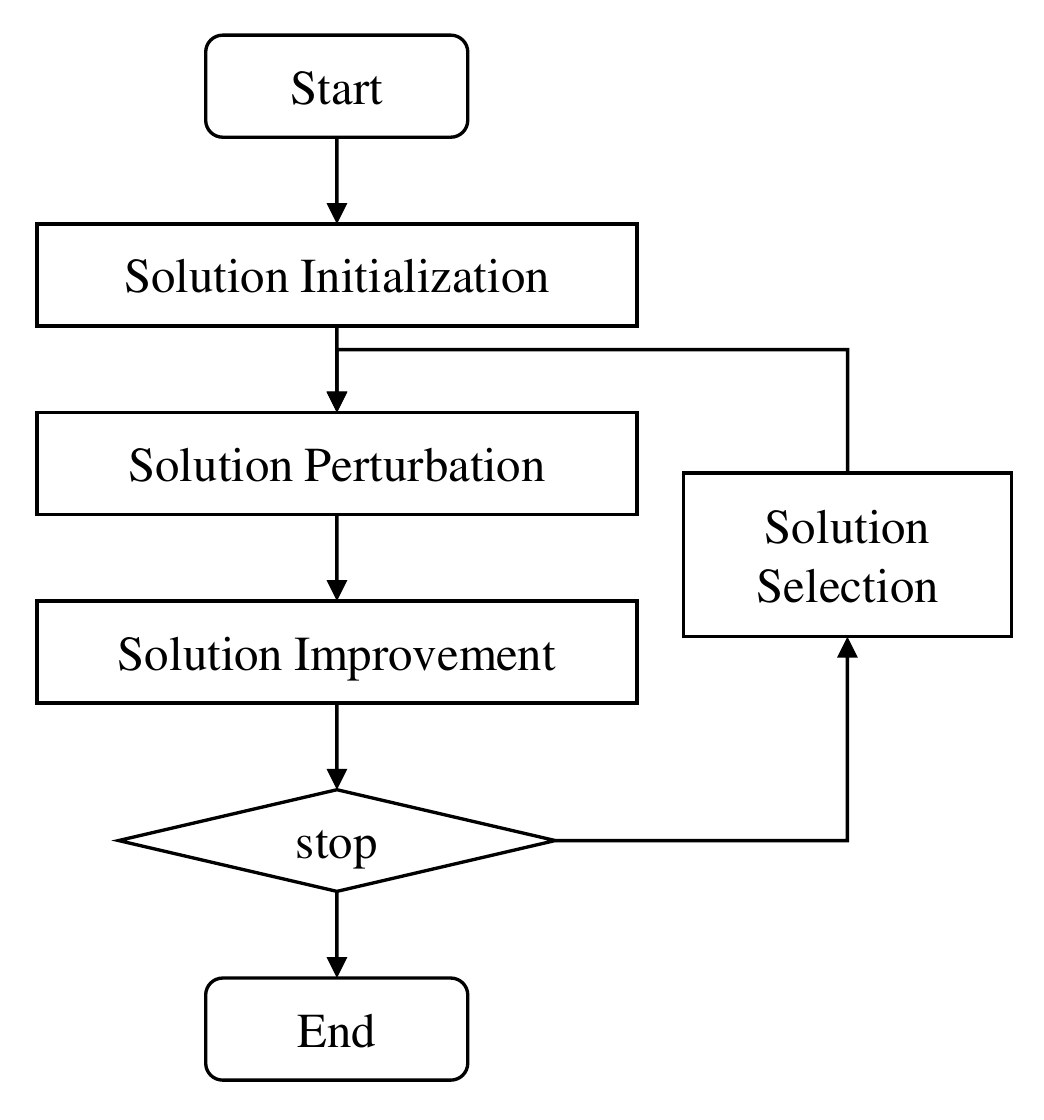}
    \caption{A general algorithm framework for SOTA heuristic methods for vehicle routing problems.}
    \label{fig:SOTA_frame}
\end{figure}

This section surveys the state-of-the-art heuristics for four widely studied vehicle routing problems: capacitated VRP, time windows VRP, multi-depot VRP, and Heterogeneous VRP~\cite{elshaer2020taxonomic}. The triumph of the cutting-edge methods benefits from a combination of several heuristic concepts rather than the implementation of a single specific metaheuristic algorithm. Despite there are various kinds of implementation, they all adhere to a general algorithm framework shown in Fig.~\ref{fig:SOTA_frame}. Besides solution initialization, there are three key components in the flow chart, which are the solution perturbation, improvement, and selection; just as their names suggest, the first one is to diversify the solution candidates, while the second is to intensify them, and the last one is to decide whether to store some new solutions or to discard some old ones.

\begin{table}[htbp]
\small
\centering
\caption{A list of state-of-the-art heuristics for four widely studied vehicle routing problems.}
\vspace{10pt}
\label{tab:SOTA_heuristics}
\renewcommand\arraystretch{1.5}
\tiny
\resizebox{1.0\textwidth}{!}{%
\begin{tabular}{p{0.1\textwidth}p{0.1\textwidth}p{0.15\textwidth}p{0.15\textwidth}p{0.15\textwidth}p{0.15\textwidth}}

    \hline
    \multicolumn{1}{c}{\textbf{Problems}} & \multicolumn{1}{c}{\textbf{Metaheuristics}}  &  \multicolumn{1}{c}{\textbf{Initialization}} &  \multicolumn{1}{c}{\textbf{Perturbation}}  &  \multicolumn{1}{c}{\textbf{Improvement}} &  \multicolumn{1}{c}{\textbf{Selection}}  \\ \hline

    \multirow{4}{*}{\shortstack{Capacitated \\ VRP}} & 
    HGS-CVRP, \cite{vidal2014unified,vidal2022hybrid} & Random    & Ordered crossover &  Relocate, Exchange, 2-opt, Insert,   Swap, 2-opt*, Swap* &  Population diversity control, Infeasible solution explore  \\ \cline{2-6} 
    
    & LNS, \cite{christiaens2020slack} & One route for each customer & Adjacent string removal, Greedy insert with   blinks & Adjacent string removal, Greedy insert with   blinks  & Simulated annealing \\ \cline{2-6} 
    
    & ILS, \cite{accorsi2021fast} & Saving method    & Shaking procedure &  2-opt, 2-opt* variants, CROSS variants   & Simulated annealing \\ \cline{2-6}
    
    & ILS, \cite{maximo2021hybrid} & Extended insert method      &  Concentric removal, Proximity removal, Sequential removal,  Proximity insert, Greedy insert  &  Relocate, Exchange, 2-opt,  Insert, Swap, 2-opt* & Solution diversity control \\ \hline
      
    VRP with Time Windows & 
    HGS-VRPTW, \cite{vidal2013hybrid,kool2022hybrid} &  Random generation, Nearest insert method, Farthest insert method, Sweep Method  &  Ordered crossover, Selective route exchange  &  Relocate, Exchange, 2-opt, Insert, Swap, 2-opt*, Swap* &  Population diversity control, Infeasible solution explore      \\ \hline
    
    \multirow{3}{*}{\shortstack{Multi-depot \\ VRP}} &
    Parallel ITS, \cite{cordeau2012parallel}  & Sweep method &  Cluster removal, Greedy insert & Tabu search & Threshold acceptance \\ \cline{2-6} 
    
    & ILS-SP, \cite{subramanian2013hybrid} & Insert method    & Multiple swap and relocate heuristics &  Relocate, Exchange, 2-opt, Insert, Swap, 2-opt* & Greedy acceptance \\ \cline{2-6} 
    
    & Unified HGS, \cite{vidal2014unified}  & Random generation &  Ordered   crossover, Assignment and insertion crossover &  2-opt, 2-opt*, CROSS, I-CROSS  &  Population diversity control, Infeasible solution explore \\ \hline

    \multirow{3}{*}{\shortstack{Heterogeneous \\ VRP}} &
    ILS-SP, \cite{subramanian2013hybrid,penna2013iterated}   &  Modified cheapest insert method, Nearest insert method &  Multiple swap and relocate heuristics, Split &  Relocate,   Exchange, 2-opt, Or-exchange, Insert, Swap, 2-opt*, K-insert & Greedy acceptance  \\ \cline{2-6} 
    
    & Unified HGS, \cite{vidal2014unified} &  Random generation  &  Ordered   crossover, Assignment and insertion crossover &  2-opt, 2-opt*, CROSS, I-CROSS  &  Population diversity control, Infeasible solution explore \\ \cline{2-6} 
    
    & HILS, \cite{penna2019hybrid}  & Inert method     &  Multiple   swap and relocate heuristics, Split, Multiple K-split, Merge &  Relocate, Exchange, 2-opt, Or-exchange, Insert, Swap, 2-opt*, K-insert  & Greedy acceptance \\ 
    \hline
\end{tabular}%
}
\end{table}

To see how SOTA methods follow the general framework in Fig.~\ref{fig:SOTA_frame}, we list in Table~\ref{tab:SOTA_heuristics} some representative ones, together with their detailed configurations with respect to different algorithm components. It should be noted that we are far from being able to review all the SOTA methods, such as the works on exact methods~\cite{baldacci2011exact,baldacci2012recent,semet2014chapter,battarra2014exact,kocc2016green,queiroga2020exact,lam2022branch} and those for routing problems with other attributes~\cite{elshaer2020taxonomic}, but by "representative" we suppose the included methods are enough to shed light on the success of SOTA methods, which we summarize as follows:

\begin{enumerate}

    \item \textit{Collaboration:} The vast majority of these methods are not simply a specific instance of some certain metaheuristic, but rather a fusion of multiple different concepts. On the one hand, effective intensification procedures, typically improvement heuristics, are combined with diversification procedures, like crossover~\cite{vidal2014unified}, shaking~\cite{accorsi2021fast}, and restart~\cite{tarantilis2013adaptive,vidal2014unified}. On the other hand, the algorithm framework may utilize multiple metaheuristic concepts to tailor the search landscape. For example, \cite{accorsi2021fast} used both iterated local search and simulated annealing.
    
    \item \textit{Initialization:} It is generally believed that a wise initialization eases the burden of the subsequent optimization process, which eventually benefits global convergence. The majority of SOTA methods achieve this by employing one or multiple constructive heuristics in the solution initialization phase. Some exceptions use simpler ways for initialization with corresponding refinement procedures. For example, in~\cite{vidal2013hybrid}, initial solutions were merely generated at random. As a remedy, it integrated powerful heuristics in solution improvement and management phases. \cite{christiaens2020slack} merely constructed one route for each customer to build the initial solution. A route-reduction strategy is designed to drive the search forward.
    
    \item  \textit{Solution Perturbation and Improvement:} Most SOTA methods combine a variety of intra-route and inter-route improvement heuristics in the solution improvement phase. Several revised improvement heuristics, such as Swap* proposed in~\cite{vidal2022hybrid}, have been shown to be efficient outside the classical ones. Compared with the improvement, the choices for perturbation are more diverse. Crossover~\cite{vidal2014unified}, shaking~\cite{kritzinger2017unified,accorsi2021fast}, restart~\cite{tarantilis2013adaptive,vidal2014unified}, among others, are typical examples. One exception is LNS~\cite{accorsi2021fast}, which employs the ruin and recreate for both perturbation and improvement. The ruin procedure removes some customers subjected to their relatedness and the recreate inserts the unassigned customers into the incumbent partial solution. Random removal, cluster removal, and string removal are typical removal heuristics used in ruin procedure. The recreate can be carried out using constructive heuristics, e.g., the insert method along with its extensions~\cite{ropke2006adaptive,pisinger2007general}, or be solved exactly as constrained programming~\cite{shaw1998using}.
    
    \item \textit{Solution Selection:} A proper management of solution(s) is vital for the long-term performance of routing heuristics. A wise solution management scheme should always avoid being short-sighted or too greedy. For the single-solution-based method, it is reflected in the designing of acceptance criteria for a new solution. Rather than accepting the new solution in a greedy way, a better strategy is to allow accepting a worse solution under a certain probability or threshold~\cite{accorsi2021fast,cordeau2012parallel}. For the population-based method, an adaptive replacement strategy is required to select the offspring generation to balance the divergence and convergence. Indeed, diversity control has been recognized as a vital component of population-based methods' competitive performance. It can be considered in various ways, such as formulated in the acquisition function~\cite{vidal2014unified} and defined in the perturbation control method~\cite{maximo2021hybrid}. The degree of diversity is typically measured by the distance between two solutions. For routing problems, edit distance~\cite{sorensen2007distance}, broken pair distance~\cite{prins2009two}, and Hamming distance~\cite{vidal2012hybrid} are among the most commonly used measurements.
    
    \item \textit{Two-phase Paradigms:} Many successful population-based heuristics for routing problems use giant tour~\cite{prins2004simple} in the representation of solution. The paradigm behind this is first to generate a giant tour and then segment it into several feasible routes. The segmentation of the giant tour can be reduced to the shortest path problem on an acyclic graph, which can be solved exactly and efficiently~\cite{vidal2016split}. This two-phase paradigm is usually referred to as the "routing first clustering second" approach~\cite{prins2014order}. Correspondingly, "clustering first routing second" is another two-phase approach, which is now commonly used in solving large-scale VRPs~\cite{comert2018cluster,costa2020cluster}. It separates customers into several clusters, then constructs a route for each cluster. The routing for each cluster can be served as a TSP variant and solved efficiently using heuristics.

\end{enumerate}

\section {Emerging Research Topics}\label{sec6}

After more than half a century, large progress has been made in designing effective heuristics and solving complex vehicle routing problems. While the majority of early works were dedicated to developing one heuristic for a specific vehicle routing problem. In the recent two decades, there has been a trend of developing more general-purpose and powerful methods. Three emerging research topics on vehicle routing heuristic can be summarized: 1) unified heuristic, 2) automatic heuristic design, and 3) machine learning-assisted heuristic.

\subsection{Unified Heuristic} 

Instead of constructing a different heuristic every time a new VRP variant was encountered, a broad consensus has been made on constructing a unified algorithm that can handle VRPs with various attributes, which is usually referred to as rich VRPs~\cite{caceres2014rich} or multi-attribute VRPs~\cite{vidal2013heuristics}. This research direction is being driven by the industry's demand that various vehicle routing scenarios can be addressed by a general-purpose routing solver to cut costs.

A thorough taxonomy for rich VRP was published by~\cite{lahyani2015rich}. Following this taxonomy, \cite{caceres2014rich} reviewed more than 50 papers concerning rich VRP. Recently, \cite{sim2019new} presented a model that generated 4800 free available new instances of rich VRPs, and provided a platform for researchers to conduct rigorous comparisons of new methods and solvers, moving academic research much closer to real practice.

Table~\ref{tab:UH_list} surveys the notable works on unified heuristics in the recent two decades along with the VRP variants they have investigated. Almost all the popular metaheuristics, including TA, LS, VNS, LNS, ALNS, and HGS, have been extended to be unified heuristics. Some of them, for example, HGS and ALNS, can handle VRPs with more than ten different attributes. Note that there are other successful unified vehicle routing solvers, which are not included here, such as the exact methods~\cite{baldacci2010exact,baldacci2011exact,markov2018unified}.

From the perspective of problem modeling formats, the various methods can be categorized into model composition and model reduction: 
\begin{enumerate}
    \item In model composition, some pre-designed templates of different attributes and constraints are stored in the algorithm. When a new problem comes, the algorithm will identify the attributes of the problem and choose the required templates to build a corresponding optimization model. For example, in ALNS~\cite{pisinger2007general}, ruin and recreate are two main procedures. The corresponding templates for various attributes of VRPs should be designed beforehand for the two procedures. HGS~\cite{vidal2013hybrid} represents the solution as a giant tour and uses an efficient split algorithm~\cite{vidal2016split} to generate the corresponding feasible solution with respect to different attributes. Crossover and local searches are performed in each iteration.  As the giant tour is a general representation, the workload of extending to rich VRPs is mainly reflected in the designing of the split algorithms and the local searches.
    \item In model reduction, all the tackled problems are transformed into one large optimization model. Generally, the large model should be complex enough to contain different attributes. In this way, the algorithm doesn't need to select templates for the new problem. The shortcoming is that the method will be less efficient when the target problem is simple and most of the attributes considered in the large model are redundant. Examples along this line, include muPOPTW~\cite{tricoire2010heuristics} and TDVRPSTW~\cite{kritzinger2017unified}. The former transformed all the target problems into a multi-period orienteering problem with multiple time windows and a variable neighborhood search method was used to solve it. The latter formulated a time-dependent vehicle routing problem with soft time windows. It used variable neighborhood search with an empirically determined parameter setting for eight VRPs and generated several new best-known results on benchmarks.
\end{enumerate}

\begin{table}[htbp]
\caption{A list of unified heuristics for rich vehicle routing problems.}
\vspace{10pt}
\label{tab:UH_list}
\centering
\footnotesize
\renewcommand\arraystretch{1.2}
\resizebox{0.9\textwidth}{!}{%
\begin{tabular}{p{0.15\textwidth}p{0.3\textwidth}p{0.6\textwidth}}
    \hline
    \multicolumn{1}{c}{\textbf{Publications}} & \multicolumn{1}{c}{\textbf{Heuristics}} & \multicolumn{1}{c}{\textbf{Applications}} \\
    \hline

    \cite{cordeau2001unified} \cite{cordeau2004improved} (2001, 2004) & Tabu search (TS) & capacitated VRP, VRP with time windows, multi-depot VRP, periodic VRP, site-dependent VRP, pickup and delivery problem, VRP with duration constraints\\
    \hline
    
    \cite{ropke2006unified} \cite{pisinger2007general} (2006,2007) & Adaptive large neighborhood search (ALNS) & capacitated VRP, VRP with time windows, multi-depot VRP,  site-dependent VRP, open VRP, VRP with backhauls, mixed VRP with backhauls, multi-depot mixed VRP with backhauls, VRP with backhauls and time windows, mixed VRP with backhauls and time windows, VRP with simultaneous deliveries and pickups \\
    \hline
    
    \cite{irnich2008unified} (2008)   & Local search (LS) & capacitated VRP, VRP with time windows, multi-depot VRP with time windows, pickup and delivery problem, periodic VRP \\
    \hline
    
    \cite{tricoire2010heuristics} (2010)   & Variable neighborhood search (VNS)   & orienteering problem, team orienteering problem, orienteering problem with time windows, team orienteering problem with time windows, multi-period orienteering problem with multiple time windows\\
    \hline

    \cite{cordeau2012parallel} (2012)   & Iterated local search (ILS) & capacitated VRP, VRP with time windows, multi-depot VRP, multi-depot VRP with time windows, periodic VRP,  periodic VRP with time windows, split delivery VRP, split delivery VRP with time windows\\
    \hline
    
    \cite{subramanian2013hybrid} (2013)  & Iterated local search with set partition (ILS-SP)     & capacitated VRP, asymmetric VRP, open VRP, VRP with simultaneous pickup and delivery, VRP with mixed pickup and delivery, multi-depot VRP, multi-depot VRP with mixed pickup and delivery\\
    \hline
    
    \cite{vidal2013hybrid} \cite{vidal2014unified} (2013, 2014)  & Hybrid   Genetic Search (HGS) & capacitated VRP, VRP with time windows, multi-depot VRP, periodic VRP, VRP with heterogeneous fleet, site-dependent VRP, VRP with profits, VRP with backhauls, asymmetric VRP, open VRP, VRP with mixed pickup and delivery, VRP with simultaneous pickup and delivery, VRP with mixed vehicle fleet, VRP with duration constraints, VRP with time windows, VRP with multiple time windows, VRP with general time windows, time-dependent VRP, VRP with flexible travel time, VRP with lunch breaks, VRP with truck-driver scheduling, generalized VRP\\
    \hline
    
    \cite{kritzinger2017unified} (2017) & Variable neighborhood search (VNS) & capacitated VRP, open VRP, open VRP with time windows, time-dependent VRP, time-dependent VRP with time windows \\
    \hline
    
    \cite{penna2019hybrid} (2019)  & Iterated local search (ILS) & capacitated VRP, VRP with time windows, multi-depot VRP, split delivery VRP, multi-trip VRP, open VRP, VRP with duration constraints, site-dependent VRP, VRP with backhauls   \\
    \hline
    
    \cite{rabbouch2021efficient} (2021) & Genetic algorithm (GA) & capacitated VRP, VRP with time windows, multi-depot VRP, VRP with heterogenous fleet, multi-depot heterogeneous VRP with time windows\\
    \hline

\end{tabular}%
}
\end{table}

Further increase of VRP variants that consider complex real-world attributes is expected in future works. The constraints can be themselves an NP-hard combinatorial optimization problem, such as 3D-bin packing~\cite{bortfeldt2020split,rajaei2021split}, which can hardly be tackled by using the existing unified frameworks. In addition, from the optimization problem modeling perspective, the VRPs can be formulated in a more general way, such as the mathematical programming language, to reduce the burden of design and implementation of the algorithm. Those general modeling languages should be tailored for vehicle routing problems. \cite{benoist2014mathematical} presented a good example, which took into consideration the formulating of sequential variables and combinatorial constraints in its unified modeling language.

\subsection{Automatic Heuristic Design}

Despite the popularity of vehicle routing heuristics, designing an efficient heuristic consumes a large amount of time, and requires extensive expert knowledge. Automatic algorithm design (AAD)~\cite{gendreau2010handbook}, also termed as hyper-heuristics~\cite{burke2013hyper,burke2019classification} or automatic design of heuristics~\cite{stutzle2019automated}, is used to tackle the challenge. It is able to automatically design a problem-tailored algorithm with the help of the knowledge learned from offline or online data. 

The works along this line of automatic heuristic design can be categorized into three levels:
\begin{enumerate}
    \item Automatic algorithm configuration focuses on automatically tuning the parameters of algorithms. Adjusting the weights of the ruin and recreate operations in ALNS~\cite{ropke2006adaptive} can be regarded as an online configuration. The online configuration can be found in many other heuristics. For example, a self-tuning heuristics, which dynamically tunes the parameters of the local search method during the optimization process, was proposed in~\cite{alabas2008self}. It was validated to be effective on the tested multiobjective routing problems. \cite{zennaki2010new} designed decision rules to predict solution quality to adaptively guide the search neighborhood of a tabu algorithm. \cite{nalepa2016adaptive} adaptively adjusted the population size, the child solution size, and the selection method in a memetic algorithm framework and satisfying results were observed on VRPTW instances. More recently, \cite{lu2019learning} used reinforcement learning to select the improvement heuristics. There are also offline approaches such as~\cite{rasku2014automating} used different tuning methods to configure the local search framework proposed in~\cite{groer2010library} and verify the advantages of automatic configuration. 
    \item Different from configuring the parameters for a fixed algorithm framework, automatic algorithm selection chooses the most suitable algorithm from a bunch of alternatives. These works typically utilize a class of classical heuristics and apply machine learning techniques to select the best algorithm. \cite{sabar2014dynamic} used an online heuristic selection technique that uses a dynamic multi-armed bandit with an extreme value-based reward to choose the best heuristic to use at each iteration. The results are demonstrated on various combinatorial optimization problems including dynamic VRP. \cite{mayer2018simulation} built both classification and regression models using artificial neural networks to select the best algorithm for dynamic VRPs. \cite{gutierrez2019selecting} used meta-learning to train a classifier to select the algorithm. It achieved an accuracy of more than 50\% in selecting the best algorithm and generated the best average performance compared to the four heuristic algorithms. \cite{jiang2021feature} created a new group of graph features to extract the hidden information. The experiments were carried out on real-world test instances and the method was used to predict the best algorithm from six heuristic algorithms. The results outperformed other algorithm selection methods for VRPs and the effectiveness of the new features was verified.
    \item Automatic algorithm composition builds the algorithm from algorithmic components rather than using existing algorithm framework~\cite{meng2021automated}. This component-based approach outperforms other approaches in the flexibility of constructing algorithms and the ability to generate novel algorithms. To our best knowledge, there is no method specially developed for VRP. Several related works can be found on other combinatorial optimization problems. For example,~\cite{lopez2012automatic} proposed an algorithm composition method based on a multiobjective ant colony algorithm for TSP. \cite{mascia2013grammars} and~\cite{franzin2019revisiting} designed iterated greedy algorithm and simulated annealing for flow shop problems. Recently, more general frameworks are developed. Typical works include a component-wise framework AutoEMOA~\cite{bezerra2015automatic} for multiobjective optimization and a general local search algorithm design framework AutoGCOP~\cite{meng2021automated} for combinatorial optimization. 
\end{enumerate}

The success of the learning process in AAD depends on the problem features. Various vehicle routing features have been used in the past decades. The most commonly-used and easy-to-generate features are the geometric features~\cite{hutter2014algorithm}. The second category of features is probing features~\cite{smith2011discovering,rasku2016feature}, which are the measurements of the behavior of some simple heuristics on the test set. The probing features reflect additional landscape information of the corresponding search heuristics compared to the basic problem features. Except for the geometric and probing features, \cite{jiang2021feature} recently presented the first attempt to use graph features, which are extracted using deep learning techniques, for automatic vehicle routing algorithm selection.

For future works, as we've mentioned, despite the existence of several general-purpose combinatorial optimization solvers~\cite{bezerra2015automatic,meng2021automated}, a composition-based automatic heuristic design method specifically for vehicle routing is still lacking. Additionally, new ways of extracting problem features are waiting to be applied. Except for the classical geometric features, deep learning~\cite{jiang2021feature} and landscape analysis techniques~\cite{malan2021survey}, which can dig deeper and optimization-related information, are some promising approaches that have been rarely studied yet.

\subsection{Machine Learning-assisted Heuristic}

Recently, the prosperity of machine learning inspires people to introduce it to the community of operational research. Along with the success of the applications in many other combinatorial optimization problems~\cite{bengio2021machine}, extensive work has been carried out in the recent two decades to integrate ML for vehicle routing. ML can be either used in an end-to-end way or to assist other algorithms. The end-to-end approach is promising but still in its infancy and can hardly beat classical heuristics~\cite{nazari2018reinforcement,kotary2021end,xu2021reinforcement}. A thorough review is out of the scope of this article. We will focus on ML-assisted vehicle routing heuristics.

Machine learning can be used to assist heuristics in various forms~\cite{bengio2021machine}. In order to classify the existing works in a principled way, we follow the taxonomy presented in the survey paper~\cite{karimi2022machine}. The ML-assisted heuristics are categorized into six classes according to the learning targets: 
\begin{itemize}
    \item Algorithm selection: select the best algorithm for the target instance from a set of algorithms.
    \item Fitness evaluation: replace the fitness function evaluation.
    \item Initialization: improve the quality or efficiency of generating initial solutions.
    \item Parameter setting: set the algorithm parameter.
    \item Evolution: assist the search process, e.g., select the local search operator.
    \item Cooperation: improve the cooperation of several heuristics.
\end{itemize}
Table~\ref{tab:ML-assisted Heuristics} lists the recent works on ML-assisted heuristics for VRPs including the heuristics they used, the learning techniques, and the applications. The observations are as follows: 1) Machine learning has been extensively applied to assist vehicle routing heuristics in the recent two decades, the works can be found in all six classes of learning targets. Capacitated VRP and VRP with time windows are the most investigated. 2) Among different heuristics, metaheuristics are most frequently used as the baseline heuristics, which fit in with the dominant position of metaheuristics in recent years. 3) The majority of existing works adopt classical machine learning techniques, such as k-means in the algorithm initialization. Meanwhile, there is a growing trend of applying deep learning, which will also be discussed in the following paragraph.

With the growing capacity and speed of computers and the development of machine learning techniques, increasing attention has been put on applying deep learning to vehicle routing heuristics. Instead of using deep learning in an end-to-end way~\cite{nazari2018reinforcement,kotary2021end,bengio2021machine,xu2021reinforcement}, many recent works start to focus on deep learning-assisted heuristics in order to beat classical heuristics. They either produce better results than classical heuristics or generate competitive results with a significantly lower time cost. For example,~\cite{chen2019learning} and~\cite{wu2021learning} adopted deep learning to design new local search operators. The results outperformed other learning-based methods and were faster than the heuristic solver. \cite{zong2022rbg} proposed a rewriting-by-generating framework for large-scale routing problems, where the rewriting was guided by reinforcement learning. \cite{santana2022neural} adopted a heatmap, which was generated by a pre-trained deep learning model, to guide the local search and crossover in HGS to enhance the performance.

\begin{table}[htbp]
\caption{A list of ML-assisted heuristics for vehicle routing problems.}
\vspace{10pt}
\centering
\renewcommand\arraystretch{1.2}
\label{tab:ML-assisted Heuristics}
\tiny
\resizebox{1.0\textwidth}{!}{%
\begin{tabular}{p{0.1\textwidth}p{0.2\textwidth}p{0.1\textwidth}p{0.2\textwidth}p{0.2\textwidth}}
    \hline
    \multicolumn{1}{c}{\textbf{Categories}}    & \multicolumn{1}{c}{\textbf{Publications}}  & \multicolumn{1}{c}{\textbf{Heuristics}}  & \multicolumn{1}{c}{\textbf{Learning Techniques}}  & \multicolumn{1}{c}{\textbf{Applications}} \\ 

    \hline

    \multirow{5}{*}{\shortstack{Algorithm\\Selection}} & \cite{de2017meta} (2017)  & GRASP, SA, LNS   &  k-nearest neighbor & VRP     \\ 
    &\cite{mayer2018simulation} (2018)  & Greedy, Re-planning   &  artificial neural network & dynamic VRP    \\ 
    &\cite{rasku2019feature} (2019) & 14 heuristics  &  nearest neighbor, multilayer perceptron, random forest, support vector machine & capacitated VRP\\ 
    &\cite{gutierrez2019selecting} (2019) & EA, GA, PSO   &  artificial neural network & VRP with time windows\\ 
    &\cite{jiang2021feature} (2021)  & TS, SA, LS, et al.    &   k-nearest neighbor, support vector machine, multilayer perceptron & capacitated VRP \\ 
    \hline
    
    \multirow{3}{*}{\shortstack{Fitness \\ Evaluation }  }  & \cite{lucas2020reducing} (2020)  & VNS   &  decision tree  & capacitated VRP  \\ 
    & \cite{costa2021learning} (2021) & GLS &  support vector machine, random forest, genetic programming & capacitated VRP \\
    & \cite{niu2022multi} (2022) & EA &  radial basic function network & stochastic VRP  \\
    \hline
    
    \multirow{5}{*}{Initialization}  & \cite{diaz2010population} (2010)  & GA    &  k-means & VRP with time windows\\ 
    & \cite{xiang2016clustering} (2016) & ACO   &  k-means & split delivery VRP\\ 
    & \cite{gocken2019comparison} (2019) & GA    &  k-means & VRP with time windows\\ 
    & \cite{min2019maximum} (2019) & TS    &  k-means & split delivery VRP\\ 
    & \cite{jingjing2022cluster} (2022) & TS    &  k-means & VRP with workload balance\\ 
    \hline
    
    \multirow{6}{*}{\shortstack{Parameter \\ Setting}}  & \cite{walker2012vehicle} (2012)  & ILS   &  credit assignment & VRP with time windows  \\ 
    & \cite{li2015iterated} (2015) & ILS   &  credit assignment & multi-depot VRP with simultaneous deliveries and pickups\\ 
    & \cite{li2016two} (2016)  & VNS   &  credit assignment & prize-collecting  periodic VRP\\ 
    & \cite{chen2016multi}   (2016)   & VNS   &  credit assignment & periodic VRP\\ 
    & \cite{peng2019memetic}  (2019)    & MA    &  credit assignment & green VRP \\ 
    & \cite{lu2019learning} (2019) & ILS   &  credit assignment & capacitated VRP  \\
    \hline
    
    \multirow{9}{*}{Evolution} & \cite{santos2006combining} (2016)    & GA    &  Apriori algorithm & oil collecting VRP  \\ 
    & \cite{al2018hybridizing} (2018) & GRASP &  Apriori algorithm & capacitated VRP  \\
    & \cite{arnold2019makes} (2019) & GLS   &  decision tree, support vector machine, random forest   & capacitated VRP  \\ 
    & \cite{arnold2021pils}   (2021)  & GA, GLS  &  Apriori algorithm & capacitated VRP  \\ 
    & \cite{xiang2021pairwise} (2021)    & ACO  &  radial basic function & dynamic VRP  \\
    & \cite{pugliese2022combining} (2022) & VNS &  reinforcement learning & VRP with crowd-shipping\\ 
    & \cite{cooray2017machine}   (2017)   & GA    &  k-means & green VRP     \\ 
    & \cite{vzunic2020adaptive}  (2020)    & TS    &  linear model, support vector machine & heterogeneous VRP with time windows   \\ 
    & \cite{ruther2022priori}   (2022)   & GA    &  Bayesian optimization & pickup and delivery problem    \\ 
    & \cite{qi2022qmoea}   (2022)   & MOEA  &  reinforcement learning & time-dependent green VRP with time windows   \\ 
    \hline
    
    \multirow{7}{*}{Cooperation} & \cite{le2005guided} (2005) & TS    &  Apriori algorithm & VRP with time windows \\ 
    & \cite{meignan2008coalition,meignan2010coalition} (2008, 2010)  & EA    &  reinforcement learning  & capacitated VRP  \\ 
    & \cite{barbucha2010cooperative} (2010) & LS    &  reinforcement learning  & VRP  \\ 
    & \cite{silva2015multi} (2015)  & ILS   &  reinforcement learning  & VRP with time windows  \\ 
    & \cite{martin2016multi} (2016)   & Saving method   &  Apriori algorithm & capacitated VRP      \\ 
    & \cite{silva2019reinforcement} (2019)   & ILS   &  reinforcement learning & VRP with time windows \\ 
    & \cite{qin2021novel} (2021)  & ACO, GA, PSO, et al.   &  reinforcement learning  & heterogeneous VRP    \\ 
    \hline
    \hline
\end{tabular}%
}
\end{table}

From our perspective, the scalability of solvers remains the major issue for ML-assisted heuristic methods. In the majority of existing works, the ML-assisted methods are trained on instances with similar distribution and attributes as the target VRP. But this assumption may no longer hold facing real-world applications, as problem instances can be obtained from any scenario and have no guarantee to be in the same distribution as the target problem. Transfer learning and multitask learning are two possible ways to perform generalization. In recent years, several applications of transfer learning and multitask learning on vehicle routing were carried out~\cite{feng2015memes,feng2019solving,feng2020towards,shang2022solving}. However, they only considered basic VRP variants. An extension to learning across VRPs with multiple attributes is a possible research direction. Except for learning across different problems, the generalization to larger vehicle routing problems is also challenging. Typically, the ML model is trained on instances with a limited size to keep the computational cost acceptable. \cite{fu2021generalize} and~\cite{zong2022rbg} broke down the large vehicle routing problem into smaller subproblems and applied the pre-trained model on each subproblem to tackle this issue.

\section{Conclusion}\label{sec7}

This article presents a systematic survey of vehicle routing heuristics. Great effort has been put into developing heuristics for solving various vehicle routing problems over the past decades. The methods are classified into three categories: constructive heuristics, improvement heuristics, and metaheuristics. Their methodologies, recent extensions, and applications are reviewed. Constructive heuristics and improvement heuristics made up the majority of early-stage methods. They are now often adopted as the algorithmic components of more advanced methods. Metaheuristics are becoming dominant, which is reflected not only in the broad research but also in their general performance on diverse vehicle routing problems. 

The current SOTA heuristics for widely studied VRP variants are summarized and analyzed. Despite their various implementations, a general algorithm framework that includes initialization, perturbation, improvement, and selection can be concluded. We discuss the different heuristics they used for each algorithm component and provide insights into the success of these methods. They are typically a hybridization of different heuristic concepts rather than an implementation of a single heuristic. The principle is to consider both search intensification and diversification through the cooperation of different well-designed heuristics.

The recent research trend is embodied in three emerging research topics: unified heuristic, automatic heuristic design, and machine learning-assisted heuristic. Unified heuristics aim at using one algorithm framework to solve a broad class of vehicle routing problems with multiple attributes. The works in the other two topics utilize historical knowledge to design or enhance classical vehicle routing heuristics. They are dedicated to developing more general-purpose and powerful heuristics for vehicle routing problems. 




\end{document}